\documentclass[lettersize,journal]{IEEEtran}
\usepackage[utf8]{inputenc}
\usepackage{amsmath,amsfonts}
\usepackage{algorithmic}
\usepackage{algorithm}
\usepackage{array}
\usepackage[
caption=false,
font=footnotesize,
labelfont=rm,
textfont=rm
]{subfig}
\usepackage{textcomp}
\usepackage{stfloats}
\usepackage{url}
\usepackage{verbatim}
\usepackage{graphicx}
\usepackage{multirow}
\usepackage{booktabs}
\usepackage{cite}
\usepackage{amsthm} 
\usepackage{lmodern}
\usepackage{newtxtext, newtxmath}
\usepackage{array}
\usepackage{booktabs}
\usepackage{multirow}
\usepackage{graphicx}
\usepackage{xcolor}
\usepackage{arydshln}
\usepackage{tabularray}
\newcommand{\removelatexerror}{\let\@latex@error\@gobble}
\begin{document}

\title{Learning Local Heterogeneity and Cross-Region Context for Large-Scale Traffic Forecasting}
\author{Qi Feng, Zidong Wang, Bo Li, Xiaoguang Gao, Jiayu Zhang, Chenfeng Wang, Kaifang Wan
\thanks{The work described in this paper was supported in part by the National Key Laboratory
	of Air-based Information Perception and Fusion (Grant No. 202471), the Innovation Foundation for Doctor Dissertation of Northwestern Polytechnical University, China (CX2026047), the National Nature Science Foundation of China (Project No. 52402453, 62003267, 62606427).(\textit{Corresponding authors:  Kaifang Wan})	
	
	Qi Feng and Jiayu Zhang are with the School of Electronic and Information, Northwestern Polytechnical University, Xi’an 710129, China. (e-mail: \{{fq19990906}, {zhangjiay}\}@mail.nwpu.edu.cn)	
	
	Bo Li, Xiaoguang Gao, and Kaifang Wan are with the School of Electronic and Information, Northwestern Polytechnical University, Xi’an 710129, China.(e-mail:\{{libo803}, {cxg2012}, {wankaifang}\}@nwpu.edu.cn)		
	
	Zidong Wang is with the Department of Computer Science, City University of Hong Kong, Hong Kong SAR, and also with Shenzhen Research Institute, City University of Hong Kong, China.(email: zidowang@cityu.edu.hk)
	
	Chenfeng Wang is with the School of Electronic Information (School of Artificial Intelligence), Northwest University, Xi'an, China. (email: wcf2026@nwu.edu.cn)}
}	
\markboth{Journal of \LaTeX\ Class Files,~Vol.~14, No.~8, August~2021}%
{Shell \MakeLowercase{\textit{et al.}}: A Sample Article Using IEEEtran.cls for IEEE Journals}
\IEEEpubid{0000--0000/00\$00.00~\copyright~2021 IEEE}
\maketitle

\begin{abstract}
Traffic flow forecasting is essential to intelligent transportation systems. Large-scale traffic forecasting requires jointly modeling local spatial dependencies and cross-region context. Spatial dependencies between geographically neighboring nodes are heterogeneous due to differences in road identity and travel direction, while acquiring global information through all-pairs node interactions incurs substantial computational costs. Therefore, capturing local heterogeneity while efficiently acquiring long-range context remains an important challenge in large-scale traffic forecasting. To address these challenges, we propose \texttt{LoReST}, a \underline{Lo}cal-\underline{Re}gion \underline{S}patial \underline{T}emporal network that models spatial dependencies at two complementary granularities: node neighborhoods and road network regions. Specifically, relation-aware local aggregation captures heterogeneous dependencies within geographic neighborhoods through road and direction specific feature transformations. Cross-region interaction constructs region representations through  mean pooling, exchanges long range context via inter-region attention, and broadcasts it back to nodes. By integrating local information aggregation with cross-region interaction, \texttt{LoReST} is able to effectively achieve spatial dependency learning in large-scale road networks. Experiments on four datasets of the LargeST benchmark show average relative reductions of 4.78\%, 3.60\%, and 5.75\% in MAE, RMSE, and MAPE, respectively.
\end{abstract}


\section{Introduction}
\IEEEPARstart{T}{raffic} flow forecasting aims to estimate future traffic flow from historical observations and is a fundamental task in intelligent transportation systems\cite{6894591}. Accurate forecasts support congestion alerts, traffic control, and route planning, helping transportation authorities identify potential bottlenecks in advance and develop appropriate control strategies\cite{zhang2008dynacas, 10208110}. As traffic sensor deployments expand, forecasting tasks increasingly extend from local roads to large-scale networks comprising thousands of nodes\cite{liu2023largest}. This broader spatial coverage not only increases the volume of data to be processed but also places greater demands on models to capture complex spatial dependencies under limited computational resources.

Traffic observations across road networks exhibit spatial correlations\cite{yu2017spatio}. The future flow at an individual node depends not only on its own historical states but may also be influenced by traffic changes at other nodes. To capture these relationships, spatial temporal graph neural networks(STGNNs) represent traffic network as graph, organize spatial information exchange through graph neural networks\cite{kipf2016semi}, and incorporate temporal modeling modules for forecasting\cite{li2017diffusion, wu2019graph, zheng2020gman, song2020spatial, guo2019attention}. However, STGNNs explicitly compute pairwise node associations, resulting in computational and memory costs that grow quadratically with the number of nodes\cite{shao2022pre}. Message passing over sparse graphs reduces the computational cost but typically requires multiple propagation steps to reach distant nodes, while repeated neighborhood aggregation may cause node representations to become increasingly similar\cite{oono2019graph}. Therefore, effectively exploiting long-range information while controlling spatial computation costs remains an important challenge in large-scale traffic forecasting.

Recent studies have explored scalable traffic forecasting from various perspectives. STID\cite{shao2022spatial} incorporates spatial and temporal identity information into a lightweight forecasting backbone and achieves competitive results without explicit inter-node information exchange. BigST\cite{han2024bigst} reduces the cost of dynamic spatial modeling through linearized global spatial convolution, whereas RPMixer\cite{yeh2024rpmixer} combines random projections and multi-layer perceptrons to construct an efficient spatial temporal forecasting architecture. Another line of research organizes interactions through spatial partitioning. In particular, PatchSTG\cite{fang2025efficient} partitions irregularly distributed sensors into spatial patches and alternates between intra-patches attention and inter-patches to learn local and global spatial dependencies. These studies demonstrate that carefully organizing spatial representations and interactions is an important approach to balancing forecasting accuracy and efficiency on large-scale road networks.

\IEEEpubidadjcol
Spatial dependency in large-scale road networks manifest at different scales. As illustrated in Fig.\ref{fig:spatial_patterns}(a), we select a group of geographically neighboring sensors and a group of geographically distant sensors from the CA road network. Fig.\ref{fig:spatial_patterns}(b) shows that neighboring sensors tend to exhibit similar traffic patterns, suggesting that geographic proximity provides a useful prior for selecting candidate neighbors in local dependency modeling. However, geographic proximity does not imply homogeneous local spatial relationships: even spatially adjacent nodes may differ in road identity and travel direction, resulting in distinct traffic association patterns\cite{gartzke2022spatial}.

Meanwhile, Fig.\ref{fig:spatial_patterns}(c) shows that geographically distant sensors may also exhibit similar traffic patterns. This suggests that traffic observations contain nonlocal statistical associations across regions in addition to local dependencies, potentially providing predictive cues beyond the immediate neighborhood. However, performing fine-grained interactions among all nodes in a large-scale road network incurs substantial computational and memory costs. The central question addressed in this work is therefore how to combine relation-aware local modeling with efficient cross-region context interaction to effectively learn spatial dependencies in large-scale road networks.

Following this rationale, we propose \texttt{LoReST}, a local–region spatial temporal forecasting network. Specifically, we first construct a fixed-size candidate neighborhood for each node using spatial prior knowledge and design a relation-aware local spatial aggregation module. Based on road identity and travel direction, this module divides candidate neighbors into three categories: same road and same direction, same road and opposite direction, and different roads. Relation-specific feature transformations are then applied to aggregate information from each category, capturing heterogeneous local spatial dependencies within geographic neighborhoods. We further organize the large-scale road network into regions based on node locations and road identities, construct compact regional representations through intra-region aggregation, and exchange cross-region context through inter-region attention. The updated regional information is broadcast back to the corresponding nodes and fused with their local representations. The main contributions of this work are summarized as follows:

\begin{figure}[!t]
	\centering
	\includegraphics[width=0.5\textwidth]{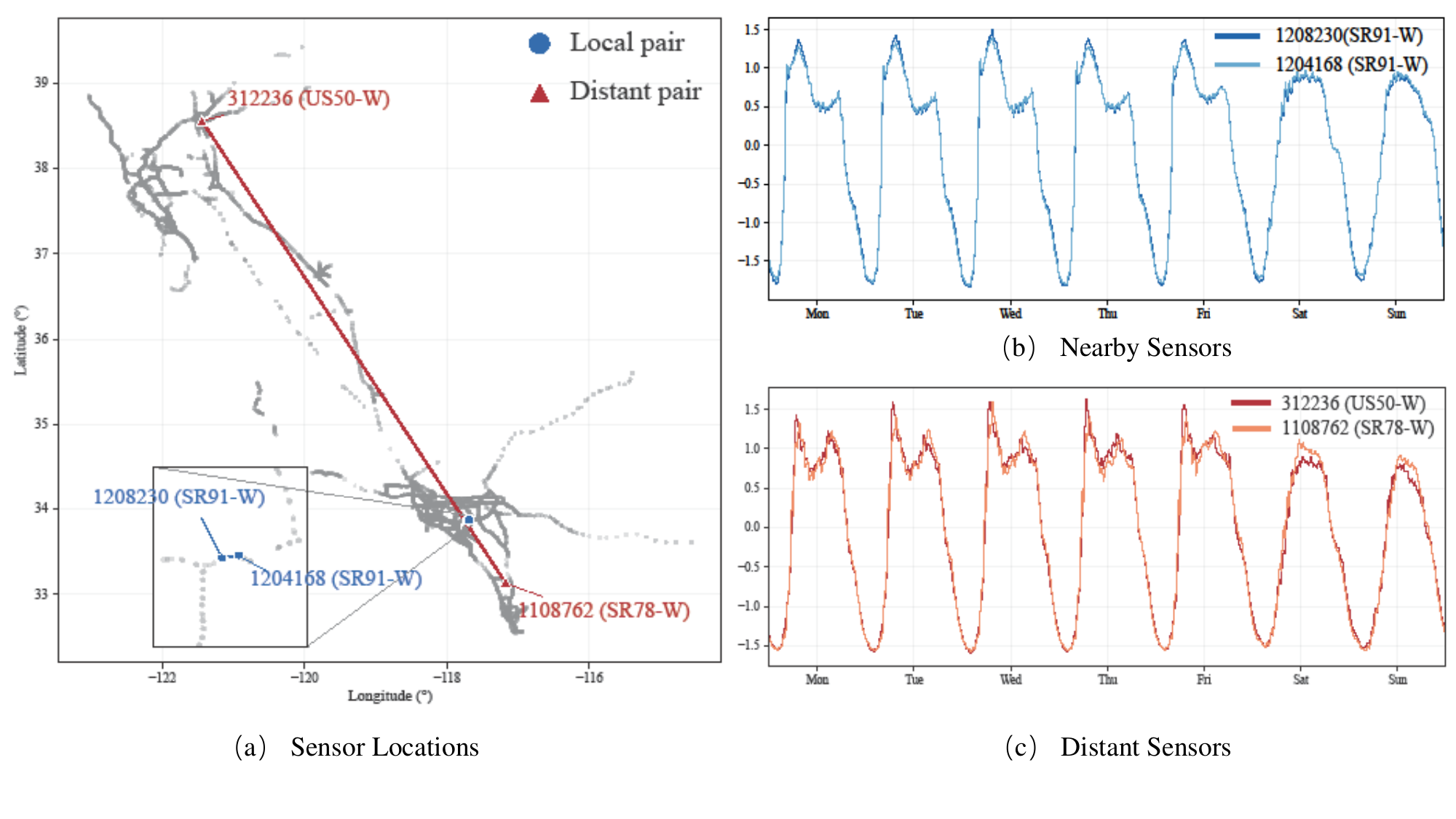}
	\caption{Examples of traffic pattern similarity at different spatial scales in the CA road network. (a) Locations of the selected sensors. (b) traffic flow of two nearby sensors. (c) traffic flow of two geographically distant sensors.}
	\label{fig:spatial_patterns}
\end{figure}

\begin{itemize}
	\item We propose \texttt{LoReST}, a local–region spatial dependency learning framework for traffic forecasting on large-scale road networks. By leveraging spatial prior knowledge, the framework organizes spatial interactions at the node and region levels to jointly model fine-grained local associations and cross-region spatial context, balancing spatial modeling capability and computational efficiency.
	\item We propose a relation-aware local aggregation mechanism to capture local heterogeneity. Using geographic proximity as a prior for candidate neighbor selection, this mechanism categorizes neighbor relations according to road identity and travel direction into same-road, same-direction; same-road, opposite-direction; and different-road relations, and performs differentiated information aggregation through relation-specific feature transformations.
	\item We propose a cross-region spatial interaction mechanism to capture long-range spatial dependencies. It constructs compact regional representations through intra-region mean pooling, exchanges spatial information over a broader range through inter-region attention, and broadcasts regional context back to individual nodes.
	\item We conduct comprehensive experiments on the real-world datasets of the LargeST benchmark. \texttt{LoReST} outperforms all compared methods on each dataset, achieving state-of-the-art forecasting performance.
\end{itemize}

The remainder of this paper is organized as follows: Section~\ref{sec: related} reviews related work on spatial dependency modeling for traffic forecasting and efficient large-scale traffic  forecasting approaches. Section~\ref{sec: preliminary} formally defines the traffic forecasting problem. Section~\ref{sec: methodology} presents the proposed framework in detail, including the model architecture, the relation-aware local aggregation, and the cross-region spatial interaction. Section~\ref{sec: experiment} reports extensive experimental results on four real-world benchmarks, accompanied by ablation studies and hyperparameter analysis. Finally, Section~\ref{sec: conclusion} concludes the paper and outlines directions for future work.

\section{Related Work}\label{sec: related}
\subsection{Spatial Dependency Modeling for Traffic Forecasting}

Traffic flow exhibits strong spatial correlations, making spatial dependency modeling an essential component of traffic flow forecasting. Since traffic nodes and their relationships can be naturally represented as graphs\cite{shao2022pre}, graph neural networks have become a mainstream approach to modeling spatial dependencies. Early spatial temporal graph neural networks (STGNNs) typically construct predefined graphs based on road connectivity or inter-node distances and aggregate neighbor information through graph convolution. For example, STGCN\cite{yu2017spatio} employs spectral graph convolution to model spatial dependencies, whereas DCRNN\cite{li2017diffusion} captures spatial information propagation through diffusion convolution based on bidirectional random walks over directed graphs. However, predefined graphs may omit relevant relational information. To reduce reliance on such graphs, GWNet\cite{wu2019graph}, AGCRN\cite{bai2020adaptive}, and MTGNN\cite{wu2020connecting} incorporate graph structure learning into end-to-end forecasting frameworks to discover latent inter-node dependencies from data. AGCRN further introduces node-adaptive parameter learning to capture node-specific patterns. However, the graph structures learned through node embeddings in these methods do not dynamically change with the input traffic conditions. Since spatial dependencies may evolve over time, fixed graph structures struggle to capture such dynamics. To address this issue, DGCRN\cite{li2023dynamic} and D2STGNN\cite{shao2022decoupled} introduce dynamic graph learning mechanisms, with D2STGNN further disentangling diffusion and inherent signals. ASTGNN\cite{9346058} uses self-attention to modulate spatial correlations and aggregates information through dynamic spatial graph convolution. DSTAGNN\cite{lan2022dstagnn} extracts spatial associations from historical traffic data and incorporates an enhanced multi-head attention mechanism to model dynamic dependencies, reducing reliance on predefined static graphs. Furthermore, PDFormer\cite{jiang2023pdformer} employs spatial self-attention with different graph masks to model short- and long-range spatial dependencies and introduces delay-aware feature transformations to capture the time delays in traffic state propagation between nodes.

Nevertheless, these methods require explicit computation or storage of dense spatial correlations, resulting in costs that grow quadratically with the number of nodes. Graph message passing requires multiple propagation steps to access long-range information. Repeated aggregation may also lead to oversmoothing, posing challenges for modeling large-scale road networks. To mitigate the over-smoothing, some studies employ neural ordinary differential equations (Neural ODEs)\cite{NEURIPS2018_69386f6b} to build continuous representations for graph neural networks, thereby overcoming the limitation of network depth and learning long-range spatial dependencies\cite{fang2021spatial,9950330}. For efficient attention-based modeling, STWave\cite{fang2023spatio} combines wavelet decomposition of traffic sequences, wavelet-based graph positional encoding, and a query sampling strategy to incorporate graph structural priors while reducing attention computation costs. Other studies explore forecasting approaches without explicit graph message passing. ST-Norm\cite{deng2021st} uses spatial and temporal normalization to highlight local and high-frequency components, respectively, enhancing the representational capacity of forecasting networks. STID\cite{shao2022spatial} addresses sample indistinguishability in the spatial and temporal dimensions by introducing spatial and temporal identity embeddings, achieving competitive forecasting performance with a lightweight multilayer perceptron backbone.

\subsection{Efficient Forecasting on Large-Scale Road Networks}

As traffic forecasting extends to large-scale road networks comprising thousands of sensors, increasing attention has been devoted to balancing computational efficiency and spatial dependency modeling capability. One line of research reduces spatial modeling costs by improving computational mechanisms. BigST\cite{han2024bigst} develops linearized global spatial convolution to learn time-varying spatial associations while enabling spatial message passing with linear complexity in the number of nodes. MAGE\cite{ma2026less} introduces a sparse and balanced mixture of adaptive graph experts, using kernel-based graph experts to learn diverse spatial relationships while maintaining linear computational complexity in the number of nodes under a fixed expert configuration. Unlike explicit graph learning approaches, RPMixer\cite{yeh2024rpmixer} adopts an all-MLP architecture and uses random projections to increase the diversity of outputs across network blocks, providing a modeling approach that does not rely on predefined spatial relationships.

Another line of research reorganizes node interactions through spatial partitioning to reduce the cost of all-node spatial computation. PatchSTG\cite{fang2025efficient} utilizes a KDtree(short for k-dimensional tree)\cite{sproull1991refinements} to partition irregularly distributed sensors and constructs equal-capacity, non-overlapping spatial patches through padding and backtracking-based merging. It then alternates between intra-patch attention and attention across nodes occupying the same index position in different patches to learn local and global spatial dependencies. SqLinear\cite{su2026efficient} further emphasizes the balance and compactness of spatial partitions, organizing sensors into near-square patches with balanced node counts. It models inter-patch dependencies and intra-patch associations through hierarchical linear interactions, reducing reliance on computationally expensive attention mechanisms.

These studies improve the scalability of large-scale traffic forecasting through advances in computational mechanisms and spatial partitioning. However, accurate forecasting requires both effective modeling of local associations around target nodes and access to long-range context beyond their immediate neighborhoods. To this end, this paper focuses on integrating local neighborhood aggregation with global context interaction for efficient and accurate large-scale traffic forecasting.

\section{Preliminary}\label{sec: preliminary}
\subsection{Traffic Flow Series}
Traffic flow series consist of observations collected by road sensors at fixed time intervals. For an individual sensor, a sequence of length \(H\) is represented as \(\mathbf{x}\in\mathbb{R}^{H}\). The observations across the entire road network are represented as \(\mathbf{X}\in\mathbb{R}^{N\times H}\), where \(N\) denotes the number of sensors, also referred to as nodes. 

\subsection{Traffic Flow Forecasting}
Traffic flow forecasting aims to predict future traffic observations from historical sequences. Given a historical observation \(\mathbf{X}_t\in\mathbb{R}^{N\times H}\), the objective is to learn a parameterized mapping \(f_\theta\) that predicts the next \(F\) time steps. In this work, sensor locations are incorporated as spatial prior information:
    \begin{equation}
    	\widehat{\mathbf{Y}}_t=f_\theta\!\left(\mathbf{X}_t,\mathbf{P}\right),
   \end{equation}
  where \(\widehat{\mathbf{Y}}_t\in\mathbb{R}^{N\times F}\) denotes the predicted traffic flow, \(\mathbf{P}\) contains the geographic prior knowledge of the sensors, and \(\theta\) denotes the learnable model parameters. Specifically,
    \[
    \mathbf{X}_t
    =
    [\mathbf{x}_{t-H+1},\ldots,\mathbf{x}_t],
    \qquad
    \mathbf{Y}_t
    =
    [\mathbf{x}_{t+1},\ldots,\mathbf{x}_{t+F}],
    \]with \(\mathbf{x}_t\in\mathbb{R}^{N}\) representing traffic observations at time step \(t\). 

\section{Methodology}\label{sec: methodology}

\begin{figure*}[!t]
\centering
\includegraphics[width=1.0\textwidth]{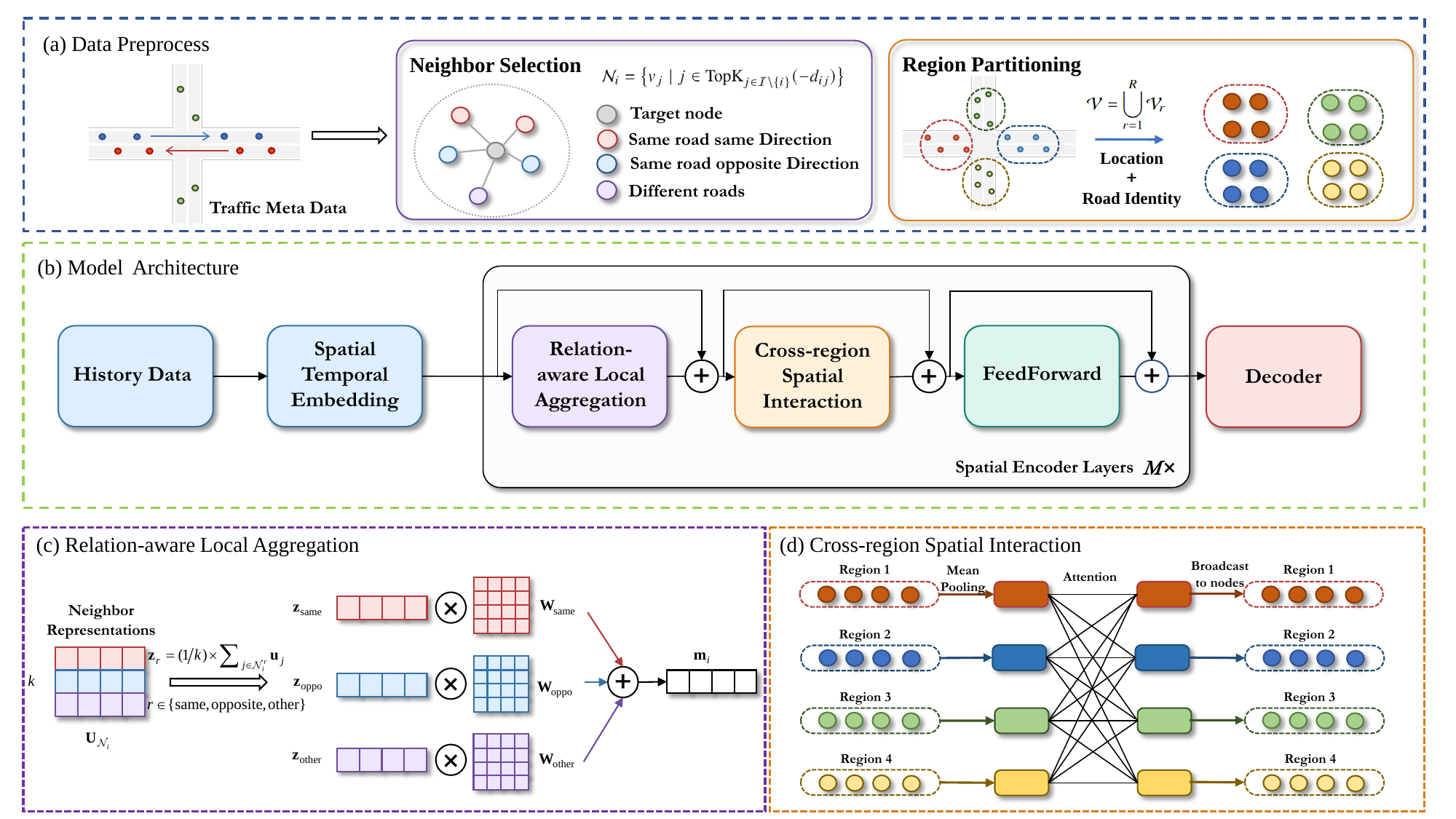}
\caption{The workflow of proposed approach. Our method comprises two stages: data preprocess and model architecture. (a) Data preprocess, we construct candidate neighborhoods and partition the road network into spatial regions based on road-network metadata, providing structural priors for local spatial aggregation and cross-region information interaction, respectively.(b) The architecture of our method. \texttt{LoReST} contains Spatial Temporal Embedding, Spatial Encoder and Decoder. The spatial temporal embedding module combines historical observations with node and time-slot identity information to construct node representations that distinguish spatial locations and temporal contexts. The spatial encoder comprises two modules: Relation-aware Local Aggregation and Cross-region Spatial Interaction, which model fine-grained local spatial dependencies and capture region-level context, respectively. Finally, the decoder maps the node representations to predictions. (c) Details of relation-aware local aggregation. (d) Details of cross-region spatial interaction.}
\label{fig: model}
\end{figure*}

The workflow of our method is illustrated in the Fig. \ref{fig: model}. This chapter provides a detailed description of the method from three aspects: overall model architecture, relation-aware local aggregation, and cross-region interaction. For clarity of presentation, neighbor selection and relation-aware local aggregation are discussed jointly, as are region partitioning and cross-region interaction.

\subsection{Model Architecture}

\subsubsection{Spatial Temporal Embedding}

Traffic observations from different sensors or time periods may exhibit similar historical patterns while corresponding to different future outcomes, making them difficult to distinguish using historical values alone. Following prior work\cite{shao2022spatial,liu2023spatio,fang2025efficient}, we incorporate spatial and temporal identity embeddings to provide node-specific and temporal contextual information. These embeddings enrich the input representations without requiring explicit pairwise interactions among all nodes.

Specifically, we first project the input traffic sequences of \(N\) nodes, \(\mathbf{X}\in\mathbb{R}^{N\times T}\), into a high-dimensional feature space through a linear transformation:
\begin{equation}
	\mathbf{E}_{\mathrm{raw}}=\mathbf{X}\mathbf{W}_{I},
\end{equation}
where $\mathbf{W}_{I} \in \mathbb{R}^{T \times D}$ is learnable parameter and \(D\) is time series embedding dimension. We then introduce two learnable temporal embeddings, \(\mathbf{E}^{\mathrm{tid}}\in\mathbb{R}^{N_d\times D_d}\) and \(\mathbf{E}^{\mathrm{diw}}\in\mathbb{R}^{N_w\times D_w}\), where \(N_d\) denotes the number of time slots per-day and \(N_w=7\) denotes the number of days per-week. They encode time-of-day and day-of-week identities, respectively, allowing the model to capture recurring temporal patterns. For each input window, the corresponding temporal indices are used to retrieve the embedding vectors. We additionally introduce a learnable spatial embeddings, \(\mathbf{E}^{s}\in\mathbb{R}^{N\times D_s}\), to represent node-specific characteristics. Finally, we concatenate the traffic representations and identity embeddings along the feature dimension to construct the initial hidden representations:
\begin{equation}
	\mathbf{H}^{(0)}=\mathbf{E}_{\mathrm{raw}}\mathbin{\|}\mathbf{E}^{\mathrm{tid}}\mathbin{\|}\mathbf{E}^{\mathrm{diw}}\mathbin{\|}\mathbf{E}^{s},
\end{equation}
where \(\|\) denotes concatenation, \(\mathbf{H}^{(0)}\in\mathbb{R}^{N\times d}\), and \(d=D+D_d+D_w+D_s\).

\subsubsection{Spatial Encoder}

Given the initial hidden representations \(\mathbf H^{(0)}\), we stack \(M\) spatial encoding layers. In the \(m\)-th layer, the Relation-aware local aggregation module (RLA) first aggregates information from candidate neighbors to obtain locally enhanced node representations \(\mathbf H_{\mathrm{loc}}^{(m)}\). The cross-region spatial interaction module (CRSI) then exchanges information among regions and propagates the resulting regional messages back to nodes, producing node representations enriched with cross-region context, \(\mathbf H_{\mathrm{cr}}^{(m)}\). Finally, a node-wise feed-forward network (FFN) further refines the features to generate the layer output \(\mathbf H^{(m)}\). Each module employs pre-layer normalization and a residual connection:
\begin{equation}
\begin{aligned}
	\mathbf H_{\mathrm{loc}}^{(m)}
	&=\mathbf H^{(m-1)}+\operatorname{RLA}\left(\operatorname{LN}(\mathbf H^{(m-1)}), \mathcal N \right),\\
	\mathbf H_{\mathrm{cr}}^{(m)}
	&=\mathbf H_{\mathrm{loc}}^{(m)}+\operatorname{CRSI}\left(\operatorname{LN}(\mathbf H_{\mathrm{loc}}^{(m)}), \mathcal V \right),\\
	\mathbf H^{(m)}
	&=\mathbf H_{\mathrm{cr}}^{(m)}+\operatorname{FFN}\left(\operatorname{LN}(\mathbf H_{\mathrm{cr}}^{(m)})\right).
\end{aligned}
\end{equation}
where $\mathcal N = [\mathcal N_1, \mathcal N_2,...,\mathcal N_N]$ is the neighbors set, $\mathcal N_i$ is the neighbor indices set of node $v_i$ and $\mathcal V = [\mathcal V_1, \mathcal V_2,...,\mathcal V_r]$ is the region indices set. Both $\mathcal N$ and $\mathcal V$ are from data preprocessing.
\subsubsection{Decoder}

Given the output of the final spatial encoding layer, \(\mathbf H^{(M)}\), we employ a linear decoder shared across all nodes to map each node’s hidden representation to traffic flow predictions for the next \(F\) time steps:
\begin{equation}
	\widehat{\mathbf Y}=\mathbf H^{(M)}\mathbf W_o+\mathbf b_o,
\end{equation}
where \(\mathbf W_o\in\mathbb R^{d\times F}\) and \(\mathbf b_o\in\mathbb R^{F}\) are learnable parameters. The prediction output is \(\widehat{\mathbf Y}\in\mathbb R^{N\times F}\). The decoder generates predictions for all \(F\) future time steps simultaneously.
\subsection{Relation-aware Local Aggregation}

\begin{figure}[!t]
	\centering
	\includegraphics[width=0.5\textwidth]{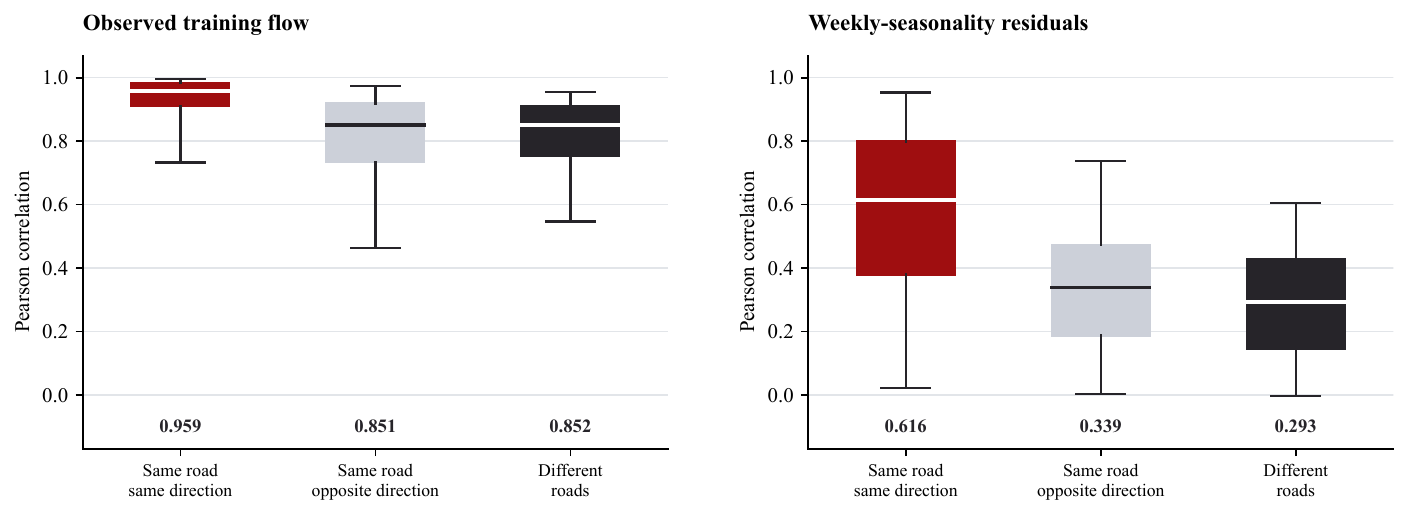}
	\caption{Traffic-flow correlation distributions across three road-relation types within selected neighborhoods (\(k=24\)) on the CA training split. Left: Pearson correlations of observed flow sequences. Right: residual correlations after removing each sensor’s training-estimated mean weekly profile.}
	\label{fig: relations}
\end{figure}

Inspired by the bidirectional diffusion modeling adopted in previous studies\cite{li2017diffusion,wu2019graph}, we incorporate road-direction information into local spatial modeling and further consider the heterogeneity of traffic associations across different road relationships.To examine these differences, we compute Pearson correlation coefficients for the three types of neighboring sensor pairs using the CA training split. As shown in Fig. \ref{fig: relations}, neighbors on the same road and in the same direction exhibit the highest correlation with the target node, significantly higher than those of neighbors on the same road but in the opposite direction and neighbors on different roads. After removing weekly seasonality, same-road same-direction neighbors still retain the highest correlation, while same-road opposite-direction neighbors show a higher correlation than different-road neighbors, and the between-group differences become more pronounced. This suggests that local spatial dependencies are not determined by geographic proximity alone, but are closely related to road topology and directional relationships. Motivated by this, we design a relation-aware neighbor aggregation module that uses relation-specific feature transformations to differentiate local neighbor information according to road identity and travel direction.

\subsubsection{Neighbor Selection}

To construct candidate neighborhoods for local spatial interaction, we measure geographic proximity using the great-circle distance defined by the Haversine formula. Let \(\mathcal I=\{1,\ldots,N\}\) denote the node index set, and let \(d_{ij}\) denote the geographic distance between nodes \(v_i\) and \(v_j\), determined from their latitude and longitude coordinates. For each target node \(v_i\), we exclude the node itself and select its \(k\) nearest neighbors:
\begin{equation}
	\mathcal N_i =\left\{v_j\mid j\in\operatorname{TopK}_{j\in\mathcal I\setminus\{i\}}(-d_{ij})
	\right\},
\end{equation}
where \(\operatorname{TopK}(\cdot)\) returns the indices of the \(k\) highest-scoring nodes. The neighbor indices are generated during preprocessing and remain fixed throughout training and inference. The local module restricts information aggregation to each target node’s candidate neighborhood, reducing the number of candidate node interactions.

\subsubsection{Relation-Aware Neighbor Aggregation}

Geographical proximity provides a useful candidate prior for modeling local dependencies among traffic nodes. To distinguish different local relationships, we design a relation-aware neighbor aggregation module that performs relation-specific message transformations.

For clarity, we present the process for an individual node. For a target node \(v_i\), we partition its candidate neighborhood \(\mathcal{N}_i\) into three mutually disjoint subsets according to the road labels of \(v_i\) and its neighbors:
\begin{equation}
	\mathcal{N}_i
	=
	\mathcal{N}_i^{\mathrm{same}}
	\cup
	\mathcal{N}_i^{\mathrm{opposite}}
	\cup
	\mathcal{N}_i^{\mathrm{other}},
\end{equation}
where \(\mathcal{N}_i^{\mathrm{same}}\), \(\mathcal{N}_i^{\mathrm{opposite}}\), and \(\mathcal{N}_i^{\mathrm{other}}\) denote neighbors on the same road with the same travel direction, neighbors on the same road with the opposite travel direction, and neighbors on different roads, respectively. Here, the direction refers to the travel direction encoded in the road label.

Given the input full node representations
\(\mathbf{H}\in\mathbb{R}^{N\times d}\), we first project them into a low dimensional feature space:
\begin{equation}
	\mathbf{U}
	=
	\mathbf{H}\mathbf{W}_{\mathrm{down}},
\end{equation}
where
\(\mathbf{W}_{\mathrm{down}}\in\mathbb{R}^{d\times d_l}\) is learnable parameter and \(\mathbf{U}=[\mathbf{u}_1,\ldots,\mathbf{u}_N]^{\top}\in\mathbb{R}^{N\times d_l}\). From the low-dimensional node representations $\mathbf{U}$, we retrieve its neighbors according to the indices in $\mathcal {N}_i$. Then the neighbor features associated with each relation type are summed separately and normalized by the total number \(k\) of candidate neighbors:
\begin{equation}
	\mathbf{z}_{i,r}
	=
	\frac{1}{k}
	\sum_{j\in\mathcal{N}_i^r}
	\mathbf{u}_j,
	\qquad
	r\in
	\left\{
	\mathrm{same},
	\mathrm{opposite},
	\mathrm{other}
	\right\}.
\end{equation}
This normalization preserves the relative proportion of each relation type within the candidate neighborhood. As illustrated in Fig.\ref{fig: model}(c), the three aggregated representations are transformed by independent relation-specific matrices and then summed to form the local message:
\begin{equation}
	\mathbf{m}_i
	=
	\sum_{r\in
		\left\{
		\mathrm{same},
		\mathrm{opposite},
		\mathrm{other}
		\right\}}
	\mathbf{z}_{i,r}\mathbf{W}_r,
\end{equation}
where
\(\mathbf{W}_r\in\mathbb{R}^{d_l\times d_l}\).
This operation allows the model to transform neighbor information differently according to the corresponding road relationship.

Finally, the fused message is projected back to the original hidden dimension \(d\) and incorporated into the node representation through a residual connection:
\begin{equation}
	\mathbf{H}_{\mathrm{loc},i}
	=
	\mathbf{H}_i
	+
	\mathbf{m}_i\mathbf{W}_{\mathrm{up}},
\end{equation}
where $\mathbf{H}_i \in \mathbb{R}^{1 \times d}$ is the input representation of target node  \(v_i\) and  \(\mathbf{W}_{\mathrm{up}}\in\mathbb{R}^{d_l\times d}\). In our implementation, we set \(d_l=D\), where \(D\) denotes the embedding dimension of the original time series. The parameters involved in this process are shared across all nodes. The neighbor aggregation for all nodes is performed in parallel via matrix operations, ultimately yielding the aggregated output representation $\mathbf{H}_{\mathrm{loc}} \in \mathbb{R}^{N \times d}$.

\subsection{Cross-Region Spatial Interaction}
Capturing spatial dependencies between distant nodes is challenging in large-scale road networks. To efficiently model long-range dependencies, we introduce a cross-region spatial interaction module.

\subsubsection{Region Partitioning}

To construct the basic units for region-level spatial interaction, we partition the road network using sensor coordinates and road identifiers. The partitioning procedure aims to maintain approximately balanced node counts across regions while promoting geographic compactness and reducing the fragmentation of nodes sharing the same road identifier. Specifically, we adopt a recursive bisection strategy to divide the node set \(\mathcal V\) into \(R\) nonempty, mutually disjoint regions:

\begin{equation}
	\mathcal V=\bigcup_{r=1}^{R}\mathcal V_r,\qquad \mathcal V_r\cap\mathcal V_{r'}=\varnothing \quad(r\ne r'),
\end{equation}
where \(|\mathcal V_r|\approx N/R\). At each bisection, we determine the node allocation ratio according to the number of regions to be formed within each child subset. This procedure is applied recursively until \(R\) regions are obtained. The resulting partition balances node counts, geographic compactness, and road affiliation consistency. It relies solely on static metadata, is also constructed during preprocessing.

\subsubsection{Cross-Region Attention}

Given the node representations produced by the local spatial aggregation module, \(\mathbf H_{\mathrm{loc}}\in\mathbb R^{N\times d}\), as shown in Fig.\ref{fig: model}(d), we construct a representation for each predefined region by averaging the representations of its constituent nodes:
\begin{equation}
	\mathbf p_r=\frac{1}{|\mathcal V_r|}\sum_{i\in\mathcal V_r}\mathbf H_{\mathrm{loc},i},
\end{equation}
where \(\mathcal V_r\) denotes the node set of region \(r\), and \(\mathbf p_r\in\mathbb R^{1\times d}\) is its regional representation. Stacking these representations row-wise yields \(\mathbf P=[\mathbf p_1^\top,\ldots,\mathbf p_R^\top]^\top\in\mathbb R^{R\times d}.\) We then apply self-attention\cite{vaswani2017attention} across regions to learn their dependencies:
\begin{equation}
	\begin{aligned}
		\mathbf Q&=\mathbf P\mathbf W_Q,\qquad \mathbf K=\mathbf P\mathbf W_K,\qquad \mathbf V=\mathbf P\mathbf W_V,\\
		\mathbf A&= \operatorname{Softmax}\left(\frac{\mathbf Q\mathbf K^\top}{\sqrt d}\right), \mathbf Z=\mathbf A\mathbf V,
	\end{aligned}
\end{equation}
where \(\mathbf W_Q,\mathbf W_K,\mathbf W_V\in\mathbb R^{d\times d}\) are learnable parameters. The matrix \(\mathbf A\in\mathbb R^{R\times R}\) contains the cross-region attention weights. The output \(\mathbf Z\in\mathbb R^{R\times d}\) contains region representations enriched with cross-region context. Finally, we linearly transform the regional outputs and broadcast to their constituent nodes through a residual update:

\begin{equation}
	\mathbf H_{\mathrm{cr},i}=\mathbf H_{\mathrm{loc},i}+(\mathbf Z\mathbf W_O)_{r(i)},
\end{equation}
where \(\mathbf W_O\in\mathbb R^{d\times d}\) is a learnable projection matrix, and \(r(i)\) denotes the region containing node \(v_i\). The resulting node representations are \(\mathbf H_{\mathrm{cr}}\in\mathbb R^{N\times d}\). Nodes within the same region receive a shared regional message while retaining their individual local representations through the residual connection.

\subsection{Model Training and Complexity Analysis}

\subsubsection{Model Training}
Given the ground-truth traffic flow \(\mathbf Y\in\mathbb R^{N\times F}\) and predictions \(\widehat{\mathbf Y}\in\mathbb R^{N\times F}\), we use the mean absolute error as the training objective:
\begin{equation}
         \mathcal L(\mathbf Y,\widehat{\mathbf Y})=\frac{1}{|\Omega|}\sum_{(i,t)\in\Omega}\left|\mathbf{Y}_{i,t}-\mathbf{\widehat Y}_{i,t}\right|,
\end{equation}
where \(\Omega\) denotes the set of node–time index pairs with valid ground-truth observations, \(N\) is the number of nodes, and \(F\) is the forecasting horizon. The model parameters are learned by minimizing $\mathcal L$ via gradient descent.
\subsubsection{Complexity Analysis}
Neighbor selection and region partitioning are performed during preprocessing and do not participate in gradient-based optimization. In low-dimensional space, constructing a KD-tree and querying the nearest neighbors of all nodes typically requires an average time complexity of \(O(N\log N+Nk)\), where \(k\) is the number of candidate neighbors. For region partitioning, assuming a fixed number of road identifiers and sorting-dominated computation, balanced recursive bisection has a recursion depth of \(O(\log R)\), yielding an upper bound of \(O(N\log N\log R)\).

The main computational components of \texttt{LoReST} are RLA and CRSI. RLA aggregates information from only \(k\) neighbors per node, resulting in a time complexity of \(O(Nk)\). In CRSI, attention is computed exclusively among regions, requiring \(O(R^2)\) operations, while region pooling and broadcasting incur an additional \(O(N)\) cost under a balanced region partition. Consequently, stacking \(M\) spatial encoder layers yields an overall forward-pass time complexity of \(O(M(Nk+N+R^2))\), which simplifies to \(O(M(Nk+R^2))\). Since \(R\ll N\) and \(k\ll N\), \texttt{LoReST} reduces the computational cost of spatial interactions compared with dense node-level attention. In particular, when \(k\), \(R\), and \(M\) are fixed, its computational complexity scales linearly with the number of nodes.
\section{Experiment}\label{sec: experiment}

To systematically evaluate \texttt{LoReST}, we conduct experimental analysis to address the following research questions:
\begin{itemize}
\item \textbf{\textit{RQ1}}: Does \texttt{LoReST} outperform existing methods across road networks of different scales and forecasting horizons?

\item \textbf{\textit{RQ2}:} How do relation-aware local aggregation and cross-region spatial interaction contribute to forecasting performance, and do they provide complementary benefits?

\item \textbf{\textit{RQ3}:} How do the key parameters affect forecasting performance?

\item \textbf{\textit{RQ4}:} Does \texttt{LoReST} learn different dependencies for different road relations?

\item \textbf{\textit{RQ5}:} Does \texttt{LoReST} capture traffic peaks, rapid transitions, and temporal trends at different forecasting horizons?
\end{itemize}
\subsection{Experiment Settings}

\subsubsection{Datasets}
We evaluate our model on four real-world traffic datasets from the LargeST benchmark: SD, GBA, GLA, and CA. Detailed dataset statistics are presented in Table \ref{tab:dataset_statistics}. We use traffic flow data from 2019 and split each dataset chronologically into training, validation, and test sets in a 6:2:2 ratio. The forecasting task uses observations from the preceding 12 consecutive time steps to predict traffic flow over the next 12 time steps.

\begin{table}[t]
	\centering
	\caption{Statistics of the LargeST datasets.}
	\label{tab:dataset_statistics}
	\resizebox{\columnwidth}{!}{%
		\begin{tabular}{lcccc}
			\toprule
			Dataset & Sensors & Time Steps & Time Interval & Timespan \\
			\midrule
			SD  & 716   & 35,040 & 15 min & 01/01/2019--12/31/2019 \\
			GBA & 2,352 & 35,040 & 15 min & 01/01/2019--12/31/2019 \\
			GLA & 3,834 & 35,040 & 15 min & 01/01/2019--12/31/2019 \\
			CA  & 8,600 & 35,040 & 15 min & 01/01/2019--12/31/2019 \\
			\bottomrule
		\end{tabular}%
	}
\end{table}

\subsubsection{Baselines}
We compare \texttt{LoReST} with representative traffic forecasting methods, including: (1) spatio-temporal graph modeling methods: AGCRN\cite{bai2020adaptive}, Graph WaveNet\cite{wu2019graph}, STGODE\cite{fang2021spatial}, DSTAGNN\cite{lan2022dstagnn}, D2STGNN\cite{shao2022decoupled}, DGCRN\cite{li2023dynamic}, and STWave\cite{fang2023spatio}; (2) lightweight MLP-based forecasting methods: STID\cite{shao2022spatial} and RPMixer\cite{yeh2024rpmixer}, where STID incorporates spatial and temporal identity embeddings and RPMixer uses random projections for feature mixing; (3) linear-complexity graph modeling methods: BigST\cite{han2024bigst} and MAGE\cite{ma2026less}; and (4) spatial partitioning-based methods for large-scale forecasting: PatchSTG\cite{fang2025efficient} and SqLinear\cite{su2026efficient}. To ensure a fair comparison, we follow official configurations of STID, MAGE, PatchSTG, and SqLinear and retrain them on our device, while the forecasting results for the remaining baselines are taken from the PatchSTG paper. 

\subsubsection{Evaluation Metrics}
To comprehensively evaluate the performance of the proposed model in traffic flow forecasting, three widely used evaluation metrics are adopted: mean absolute error (MAE), root mean squared error (RMSE), and mean absolute percentage error (MAPE). These metrics measure the deviation between the prediction and ground truth in terms of absolute error, sensitivity to large errors, and relative percentage error, respectively. For all three metrics, lower values indicate better predictive performance.

\subsubsection{Implementation Details}

We implement \texttt{LoReST} on BasicTS \cite{10726722}, an open-source time series forecasting platform. For training, we employ the AdamW optimizer\cite{DBLP:conf/iclr/LoshchilovH19} with an initial learning rate of 0.002 and a weight decay of 0.001. The model is trained for 50 epochs with a batch size of 16. During training, we adopt a fixed-epoch learning rate decay schedule, in which the learning rate is halved at epochs 2, 20, 25, and 35. For model configuration, we stack 4 spatial encoder layers. The spatial embedding dimension ${{D}_{s}}$ is 48, and the two temporal embedding ${{D}_{d}}$ and ${{D}_{w}}$ dimensions are both 24. The number of regions $R$ is set to 16. The sequence embedding dimension $D$ is set to 64, 128, 64, and 64 on SD, GBA, GLA, and CA, respectively, and the number of neighbors $k$ is set to 24, 12, 24, and 24, respectively. All experiments are conducted on an NVIDIA RTX 4090 GPU with 24 GB memory, and the random seed is fixed to 2026 for all experiments.

\subsection{Main Results(\textbf{RQ1})}

As shown in Table \ref{tab:lorest_comparison},\texttt{LoReST} achieves the best results across all evaluated prediction horizons and average metrics on the four datasets. Compared with the best-performing baseline for each metric, \texttt{LoReST} reduces average MAE by 6.74\%, 4.15\%, 4.88\%, and 3.33\% on SD, GBA, GLA, and CA, respectively. The corresponding mean relative reductions across MAE, RMSE, and MAPE are 6.25\%, 3.84\%, 5.42\%, and 3.56\%. These consistent improvements demonstrate that\texttt{LoReST} achieves state-of-the-art forecasting performance among the evaluated methods.

Graph neural networks struggle to effectively capture global dependencies on large-scale road networks, which consequently undermines the forecasting performance of STGNNs. Although dynamic graph-based methods, such as DGCRN and D2STGNN, achieve relatively outstanding performance by learning dynamic dependencies among nodes, they remain constrained by the over-smoothing problem and thus still fail to adequately learn the spatial dependencies of large-scale road networks. Similarly, while sparse graph propagation methods such as BigST and MAGE improve model scalability, they still struggle to capture complex dependencies. Notably, STWave learns spatial dependencies through a sparse graph attention mechanism and still achieves performance comparable to graph convolution methods on large-scale datasets. This demonstrates that effective forecasting does not necessarily rely on full-node interactions, and that \textbf{properly sparsified information exchange can likewise capture spatial correlations}. In contrast, STID leverages spatial-temporal identity information and achieves competitive forecasting performance without explicit spatial modeling, \textbf{highlighting the importance of identity information in traffic forecasting}. Building upon identity information, PatchSTG further devises an irregular node-partitioning scheme and employs a dual-branch attention mechanism to learn local and global information, thereby attaining remarkable forecasting performance. \textbf{This also underscores the significance of explicit spatial modeling in large-scale traffic forecasting}.

In summary, the comparative experiments demonstrate that \texttt{LoReST} consistently achieves superior predictive performance across road networks of varying scales and different forecasting horizons. Its core principle is to employ tailored interaction mechanisms at complementary spatial granularities: at the node level, road-relation-aware neighbor aggregation captures the heterogeneity of local spatial dependencies; at the region level, compact region representations facilitate the exchange of long-range context. This design preserves fine-grained local information while avoiding the substantial computational cost of all-pairs node interactions, highlighting the effectiveness of integrating spatial prior knowledge with multi-granular spatial modeling.

\begin{table*}[p]
	\centering
	\caption{Forecasting performance on the SD, GBA, GLA, and CA datasets. }
	\label{tab:lorest_comparison}
	\begingroup
	\footnotesize
	\newcommand{\lorestRed}[1]{\textbf{#1}}
	\newcommand{\lorestBlue}[1]{\underline{#1}}
	\resizebox{\textwidth}{!}{%
		\begin{tblr}{
				colspec = {Q[c,m] Q[c,m] *{12}{Q[c,m]}},
				cells = {halign=c,valign=m},
				rows = {abovesep=1.8pt,belowsep=1.8pt},
				colsep = 3pt,
				stretch = 0,
				vspan = even,
				vline{2,3} = {0.4pt},
				hline{1,Z} = {0.8pt},
				hline{2} = {3-14}{0.3pt},
				hline{3} = {0.5pt},
			}
			\SetCell[r=2]{c,m} Dataset & \SetCell[r=2]{c,m} Model
			& \SetCell[c=3]{c,m} Horizon 3 & &
			& \SetCell[c=3]{c,m} Horizon 6 & &
			& \SetCell[c=3]{c,m} Horizon 12 & &
			& \SetCell[c=3]{c,m} Average & & \\
			& & MAE & RMSE & MAPE(\%)
			& MAE & RMSE & MAPE(\%)
			& MAE & RMSE & MAPE(\%)
			& MAE & RMSE & MAPE(\%) \\
			\SetCell[r=14]{c,m} SD & GWNet & 15.15 & 25.29 & 9.82 & 17.95 & 30.39 & 11.93 & 21.82 & 38.63 & 15.09 & 17.86 & 31.00 & 11.94 \\
			& AGCRN & 15.24 & 25.13 & 9.86 & 17.74 & 29.51 & 11.70 & 21.56 & 36.82 & 15.13 & 17.74 & 29.62 & 11.88 \\
			& STGODE & 15.71 & 27.85 & 11.48 & 18.06 & 31.51 & 13.06 & 21.86 & 39.44 & 16.52 & 18.09 & 32.01 & 13.28 \\
			\SetHline{2-14}{dashed,0.35pt}
			& DSTAGNN & 18.54 & 30.33 & 11.81 & 24.55 & 40.04 & 16.51 & 35.90 & 58.31 & 27.67 & 25.25 & 42.56 & 17.64 \\
			& D2STGNN & 18.13 & 28.96 & 11.38 & 21.71 & 34.44 & 13.93 & 27.51 & 43.95 & 19.34 & 21.82 & 34.68 & 14.40 \\
			& DGCRN & 14.92 & 24.95 & 9.56 & 17.52 & \lorestBlue{29.24} & \lorestBlue{11.36} & 22.62 & 37.14 & 14.86 & 17.85 & \lorestBlue{29.51} & 11.54 \\
			& STWave & 15.34 & 25.35 & 10.01 & 18.05 & 30.06 & 11.90 & 22.06 & 37.51 & 15.27 & 18.02 & 30.09 & 12.07 \\
			\SetHline{2-14}{dashed,0.35pt}
			& RPMixer & 16.75 & 28.04 & 11.00 & 19.71 & 33.56 & 13.16 & 23.67 & 42.12 & 16.58 & 19.55 & 33.57 & 13.22 \\
			&\underline{ STID} & 15.02 & 25.40 & 10.05 & 17.95 & 30.51 & 12.16 & 21.94 & 39.44 & 15.31 & 17.91 & 31.32 & 12.14 \\
			\SetHline{2-14}{dashed,0.35pt}
			& BigST & 15.80 & 25.89 & 10.34 & 18.18 & 30.03 & 11.96 & 21.98 & 36.99 & 15.30 & 18.22 & 30.12 & 12.20 \\
			& \underline{MAGE} & 15.88 & 26.02 & 10.43 & 18.54 & 30.68 & 12.53 & 22.69 & 38.90 & 16.38 & 18.50 & 31.30 & 12.74 \\
			\SetHline{2-14}{dashed,0.35pt}
			& \underline{PatchSTG }& \lorestBlue{14.57} & \lorestBlue{24.29} & \lorestBlue{9.29} & \lorestBlue{17.43} & 29.30 & 11.43 & \lorestBlue{21.35} & \lorestBlue{36.81} & \lorestBlue{14.78} & \lorestBlue{17.35} & 29.75 & \lorestBlue{11.43} \\
			& \underline{SqLinear} & 15.40 & 25.60 & 10.15 & 18.11 & 30.38 & 12.11 & 21.52 & 37.66 & 15.04 & 17.90 & 30.76 & 12.07 \\
			\SetHline{2-14}{dashed,0.35pt}
			& \texttt{LoReST} & \lorestRed{13.94} & \lorestRed{23.41} & \lorestRed{8.96} & \lorestRed{16.14} & \lorestRed{27.39} & \lorestRed{10.57} & \lorestRed{19.66} & \lorestRed{34.11} & \lorestRed{13.49} & \lorestRed{16.18} & \lorestRed{27.85} & \lorestRed{10.70} \\
			\SetHline{2-14}{dashed,0.35pt}
			& \textbf{Improve. (\%)}
			& +4.32 & +3.62 & +3.55
			& +7.40 & +6.33 & +6.95
			& +7.92 & +7.33 & +8.73
			& +6.74 & +5.63 & +6.39 \\
			\SetHline{1-14}{0.5pt}
			\SetCell[r=14]{c,m} GBA & GWNet & 17.85 & 29.12 & 13.92 & 21.11 & 33.69 & 17.79 & 25.58 & 40.19 & 23.48 & 20.91 & \lorestBlue{33.41} & 17.66 \\
			& AGCRN & 18.31 & 30.24 & 14.27 & 21.27 & 34.72 & 16.89 & 24.85 & 40.18 & 20.80 & 21.01 & 34.25 & 16.90 \\
			& STGODE & 18.84 & 30.51 & 15.43 & 22.04 & 35.61 & 18.42 & 26.22 & 42.90 & 22.83 & 21.79 & 35.37 & 18.26 \\
			\SetHline{2-14}{dashed,0.35pt}
			& DSTAGNN & 19.73 & 31.39 & 15.42 & 24.21 & 37.70 & 20.99 & 30.12 & 46.40 & 28.16 & 23.82 & 37.29 & 20.16 \\
			& D2STGNN & 17.54 & \lorestBlue{28.94} & \lorestBlue{12.12} & 20.92 & \lorestBlue{33.92} & \lorestBlue{14.89} & 25.48 & 40.99 & 19.83 & 20.71 & 33.65 & \lorestBlue{15.04} \\
			& DGCRN & 18.02 & 29.49 & 14.13 & 21.08 & 34.03 & 16.94 & 25.25 & 40.63 & 21.15 & 20.91 & 33.83 & 16.88 \\
			& STWave & 17.95 & 29.42 & 13.01 & 20.99 & 34.01 & 15.62 & 24.96 & \lorestBlue{40.31} & 20.08 & 20.81 & 33.77 & 15.76 \\
			\SetHline{2-14}{dashed,0.35pt}
			& RPMixer & 20.31 & 33.34 & 15.64 & 26.95 & 44.02 & 22.75 & 39.66 & 66.44 & 37.35 & 27.77 & 47.72 & 23.87 \\
			& \underline{STID} & \lorestBlue{17.37} & 29.30 & 13.44 & \lorestBlue{20.47} & 34.40 & 16.17 & \lorestBlue{24.43} & 41.42 & 20.14 & \lorestBlue{20.25} & 34.58 & 16.05 \\
			\SetHline{2-14}{dashed,0.35pt}
			& BigST & 18.70 & 30.27 & 15.55 & 22.21 & 35.33 & 18.54 & 26.98 & 42.73 & 23.68 & 21.95 & 35.54 & 18.50 \\
			& \underline{MAGE} & 18.20 & 29.78 & 15.36 & 21.54 & 34.92 & 19.11 & 25.88 & 42.60 & 22.77 & 21.27 & 35.04 & 18.61 \\
			\SetHline{2-14}{dashed,0.35pt}
			& \underline{PatchSTG} & 17.68 & 29.98 & 12.73 & 20.83 & 34.74 & 15.48 & 24.94 & 41.18 & 19.64 & 20.59 & 34.73 & 15.51 \\
			& \underline{SqLinear} & 17.89 & 29.69 & 13.88 & 20.92 & 34.38 & 16.68 & 23.95 & 39.61 & \lorestBlue{19.51} & 20.46 & 34.12 & 16.25 \\
			\SetHline{2-14}{dashed,0.35pt}
			&\texttt{LoReST} & \lorestRed{16.89} & \lorestRed{28.82} & \lorestRed{11.91} & \lorestRed{19.72} & \lorestRed{33.15} & \lorestRed{14.64} & \lorestRed{23.04} & \lorestRed{38.37} & \lorestRed{17.63} & \lorestRed{19.41} & \lorestRed{32.98} & \lorestRed{14.23} \\
			\SetHline{2-14}{dashed,0.35pt}
			& \textbf{Improve. (\%)}
			& +2.76 & +0.41 & +1.73
			& +3.66 & +2.27 & +1.68
			& +3.80 & +4.81 & +9.64
			& +4.15 & +1.99 & +5.39 \\
			\SetHline{1-14}{0.5pt}
			\SetCell[r=12]{c,m} GLA & GWNet & 17.28 & 27.68 & 10.18 & 21.31 & 33.70 & 13.02 & 26.99 & 42.51 & 17.64 & 21.20 & 33.58 & 13.18 \\
			& AGCRN & 17.27 & 29.70 & 10.78 & 20.38 & 34.82 & 12.70 & 24.59 & 42.59 & 16.03 & 20.25 & 34.84 & 12.87 \\
			& STGODE & 18.10 & 30.02 & 11.18 & 21.71 & 36.46 & 13.64 & 26.45 & 45.09 & 17.60 & 21.49 & 36.14 & 13.72 \\
			\SetHline{2-14}{dashed,0.35pt}
			& DSTAGNN & 19.49 & 31.08 & 11.50 & 24.27 & 38.43 & 15.24 & 30.92 & 48.52 & 20.45 & 24.13 & 38.15 & 15.07 \\
			& STWave & 17.48 & 28.05 & 10.06 & 21.08 & 33.58 & 12.56 & 25.82 & 41.28 & 16.51 & 20.96 & 33.48 & 12.70 \\
			\SetHline{2-14}{dashed,0.35pt}
			& RPMixer & 19.94 & 32.54 & 11.53 & 27.10 & 44.87 & 16.58 & 40.13 & 69.11 & 27.93 & 27.87 & 48.96 & 17.66 \\
			& \underline{STID} & 16.39 & 27.37 & 9.72 & 19.72 & 33.37 & 12.21 & 24.25 & 42.07 & 16.07 & 19.59 & 33.91 & 12.30 \\
			\SetHline{2-14}{dashed,0.35pt}
			& BigST & 18.38 & 29.40 & 11.68 & 22.22 & 35.53 & 14.48 & 27.98 & 44.74 & 19.65 & 22.08 & 36.00 & 14.57 \\
			& \underline{MAGE} & 17.34 & 28.12 & 11.04 & 20.89 & 33.97 & 13.78 & 25.69 & 42.40 & 18.02 & 20.70 & 34.39 & 13.73 \\
			\SetHline{2-14}{dashed,0.35pt}
			&\underline{ PatchSTG} & \lorestBlue{16.08} & \lorestBlue{26.61} & \lorestBlue{9.46} & \lorestBlue{19.39} & \lorestBlue{32.26} & \lorestBlue{11.64} & \lorestBlue{23.87} & \lorestBlue{40.69} & \lorestBlue{15.19} & \lorestBlue{19.26} & \lorestBlue{32.81} & \lorestBlue{11.70} \\
			& \underline{SqLinear} & 17.16 & 28.23 & 10.30 & 20.59 & 34.34 & 12.41 & 24.86 & 41.95 & 16.51 & 20.36 & 34.51 & 12.74 \\
			\SetHline{2-14}{dashed,0.35pt}
			& \texttt{LoReST} & \lorestRed{15.57} & \lorestRed{26.01} & \lorestRed{9.09} & \lorestRed{18.49} & \lorestRed{32.13} & \lorestRed{10.85} 
			& \lorestRed{22.24} & \lorestRed{37.86} & \lorestRed{13.69} & \lorestRed{18.32} & \lorestRed{31.32} & \lorestRed{10.90} \\
			\SetHline{2-14}{dashed,0.35pt}
			& Improve. (\%)
			&+3.17& +2.25 & +3.91
			& +4.64&+ 0.40 & +6.79
			& +6.83 &+6.96 & +9.87
			& +4.88&+4.54 & +6.84 \\
			\SetHline{1-14}{0.5pt}
			\SetCell[r=10]{c,m} CA & GWNet & 17.14 & 27.81 & 12.62 & 21.68 & 34.16 & 17.14 & 28.58 & 44.13 & 24.24 & 21.72 & 34.20 & 17.40 \\
			& STGODE & 17.57 & 29.91 & 13.91 & 20.98 & 36.62 & 16.88 & 25.46 & 45.99 & 21.00 & 20.77 & 36.60 & 16.80 \\
			\SetHline{2-14}{dashed,0.35pt}
			& STWave & 16.77 & 26.98 & 12.20 & 18.97 & 30.69 & 14.40 & 25.36 & 38.77 & 19.01 & 19.69 & 31.58 & 14.58 \\
			\SetHline{2-14}{dashed,0.35pt}
			& RPMixer & 18.18 & 30.49 & 12.86 & 24.33 & 41.38 & 18.34 & 35.74 & 62.12 & 30.38 & 25.07 & 44.75 & 19.47 \\
			& \underline{STID} & 15.53 & 26.27 & 11.39 & 18.54 & 31.67 & 13.78 & 22.87 & 39.48 & 17.62 & 18.39 & 32.10 & 13.82 \\
			\SetHline{2-14}{dashed,0.35pt}
			& BigST & 17.15 & 27.92 & 13.03 & 20.44 & 33.16 & 15.87 & 25.49 & 41.09 & 20.97 & 20.32 & 33.45 & 15.91 \\
			& \underline{MAGE} & 16.54 & 27.08 & 19.81 & 19.81 & 32.36 & 15.45 & 24.40 & 40.27 & 19.91 & 19.67 & 32.76 & 15.31 \\
			\SetHline{2-14}{dashed,0.35pt}
			& \underline{PatchSTG} & \lorestBlue{14.91} & \lorestBlue{25.25} & \lorestBlue{10.26} & \lorestBlue{17.81} & \lorestBlue{30.10} & \lorestBlue{12.58} & \lorestBlue{21.73} & \lorestBlue{36.86} & \lorestBlue{16.53} & \lorestBlue{17.71} & \lorestBlue{30.37} & \lorestBlue{12.74} \\
			& \underline{SqLinear} & 15.75 & 26.42 & 11.38 & 18.68 & 31.48 & 13.81 & 22.30 &37.61  & 17.19 & 18.42 & 31.44 & 13.64 \\
			\SetHline{2-14}{dashed,0.35pt}
			&\texttt{LoReST} & \lorestRed{14.67} & \lorestRed{24.95} & \lorestRed{10.12} & \lorestRed{17.27} & \lorestRed{29.37} & \lorestRed{12.18} & \lorestRed{20.67} & \lorestRed{35.33} & \lorestRed{15.28} & \lorestRed{17.12} & \lorestRed{29.47} & \lorestRed{12.19} \\
			\SetHline{2-14}{dashed,0.35pt}
			& \textbf{Improve. (\%)}
			& +1.61 & +1.19 & +1.36
			& +3.03 & +2.43 &+3.18
			& +4.88& +4.15 & +7.50
			&+3.33 & +2.96 &+4.40 \\
		\end{tblr}%
	}
	\endgroup
\end{table*}

To ensure a fair comparison of computational efficiency, we evaluate the parameter counts, training times, and inference times of all models on the CA dataset under identical hardware and experimental settings. As shown in Table \ref{tab:efficiency}, \texttt{LoReST} achieves the lowest MAE with 1.790 million parameters. Compared with PatchSTG, \texttt{LoReST} has slightly fewer parameters and reduces training and inference times by approximately 49.2\% and 44.4\%, respectively, while further improving forecasting accuracy. Compared with SqLinear, \texttt{LoReST} has more parameters but requires only approximately 5.3\% more training time and achieves comparable inference time, while reducing MAE by approximately 7.1\%. MAGE achieves the fastest training and inference speeds, but has substantially more parameters and a higher forecasting error. Overall, \textbf{\texttt{LoReST} delivers superior forecasting performance while maintaining relatively low computational overhead, demonstrating the favorable accuracy–efficiency trade-off of its architectural design.}

\begin{table}[t]
	\caption{Comparison of Model Size, Computational Efficiency,
		and Forecasting Accuracy}
	\label{tab:efficiency}
	\centering
	\footnotesize
	\renewcommand{\arraystretch}{1.15}
	\setlength{\tabcolsep}{3pt}
	\begin{tabular*}{\columnwidth}{@{\extracolsep{\fill}}lcccc@{}}
		\toprule
		Model
		& \shortstack{Parameters(M)}
		& \shortstack{Training(s/epoch)}
		& \shortstack{Inference(s)}
		& MAE \\
		\midrule
		MAGE
		& 54.611 & \textbf{154} & \textbf{23.28} & 19.67 \\
		PatchSTG
		& 1.835 & 394 & 53.89 & 17.71 \\
		SqLinear
		& \textbf{0.242} & 190 & 30.02 & 18.42 \\
		\texttt{LoReST}
		& 1.790 & 200 & 29.96 & \textbf{17.12} \\
		\bottomrule
	\end{tabular*}
\end{table}

\subsection{Ablation Study(\textbf{RQ2})}

To evaluate the contributions of individual components and key mechanisms to forecasting performance, we design the following variants of \texttt{LoReST}:
\begin{itemize}
	\item \textbf{w/o Local:} Removes the Relation-aware Local Aggregation (RLA) module and obtains spatial context solely through cross-region spatial interaction.
	
	\item \textbf{w/o Region:} Removes the Cross-Region Spatial Interaction (CRSI) module and models spatial dependencies solely through local neighbor message aggregation.
	
	\item \textbf{FFN-only:} Removes both RLA and CRSI while retaining the spatial temporal identity embeddings and feed-forward networks, without explicit spatial information exchange between nodes.
	
	\item \textbf{w/o relation:} Shares a single feature transformation matrix across all three neighbor categories, eliminating relation-specific message transformations based on road identity and travel direction.
	
	\item \textbf{s-d only:} Allows only same-road, same-direction neighbors to participate in local message aggregation, excluding information from the other two categories.
	
	\item \textbf{w/o mean pooling:} Removes the intra-region mean pooling operation in CRSI to evaluate its effect on region representation construction and forecasting performance. Regions are padded to a uniform number of node slots, with an additive mask applied to empty slots.
	
	\item \textbf{w/ dual attention:} Replaces RLA and CRSI with intra-region and inter-region attention mechanisms, respectively, to compare alternative ways of modeling local and cross-region interactions.
\end{itemize}

The ablation results are presented in the TABLE \ref{tab:ablation}, from which we draw the following observations:

\textbf{Effectiveness of RLA and CRSI.} Removing either RLA or CRSI degrades forecasting performance, while removing both leads to further deterioration in the \textbf{FFN-only} variant. These results indicate that identity embeddings alone are insufficient to fully capture spatial dependencies between traffic nodes. Local neighbor information and cross-region context provide complementary predictive cues, and jointly modeling them helps obtain more comprehensive spatial representations. Moreover, removing CRSI causes larger increases in MAE and RMSE than removing RLA across all four datasets, highlighting the importance of cross-region information exchange in the forecasting task.

\textbf{Effectiveness of local heterogeneity learning.} Replacing the relation-specific feature transformations with a shared matrix degrades performance across all four datasets. This suggests that a uniform message mapping may struggle to accommodate local dependencies associated with different road relations. Meanwhile, although same-road, same-direction neighbors generally exhibit stronger traffic correlations, the \textbf{s-d only} variant still underperforms the complete model. This suggests that opposite-direction neighbors and neighbors on different roads can still provide complementary information. Local modeling therefore benefits from retaining multiple neighbor types and processing them according to their road relations. Notably, the \textbf{s-d only} and \textbf{w/o relation} variants even underperform the \textbf{w/o Local} variant on some datasets, indicating that local aggregation under missing or mixed information conditions can negatively contribute to forecasting.

\textbf{Effectiveness of regional representations and spatial interaction mechanisms.} Removing intra-region mean pooling degrades all forecasting metrics across the four datasets, supporting the effectiveness of compact regional representations for cross-region information exchange. This result indicates that preserving more node-level interaction details does not necessarily improve forecasting performance within our architecture. Furthermore, replacing the two modules with intra-region and inter-region attention also reduces performance, suggesting that the dual-attention alternative is less effective than \texttt{LoReST}’s combination of relation-aware local aggregation and region context interaction.

\begin{table*}[t]
	\centering
	\caption{Ablation results on the four LargeST datasets.
		Lower values indicate better performance.
		The best results are highlighted in bold.}
	\label{tab:ablation}
	\renewcommand{\arraystretch}{1.15}
	\setlength{\tabcolsep}{5pt}
	\resizebox{\textwidth}{!}{%
		\begin{tabular}{l *{12}{c}}
			\toprule
			\multirow{2}{*}{\textbf{Variant}}
			& \multicolumn{3}{c}{SD}
			& \multicolumn{3}{c}{GBA}
			& \multicolumn{3}{c}{GLA}
			& \multicolumn{3}{c}{CA} \\
			\cmidrule(lr){2-4}
			\cmidrule(lr){5-7}
			\cmidrule(lr){8-10}
			\cmidrule(lr){11-13}
			& MAE & RMSE & MAPE (\%)
			& MAE & RMSE & MAPE (\%)
			& MAE & RMSE & MAPE (\%)
			& MAE & RMSE & MAPE (\%) \\
			\midrule
			
			\textbf{w/o Local}
			& 16.49 & 28.44 & 10.80
			& 19.43 & 33.09 & 14.36
			& 18.79 & 32.17 & 11.45
			& 17.58 & 30.39 & 12.61 \\
			
			\textbf{w/o Region}
			& 17.26 & 30.72 & 11.18
			& 19.70 & 33.90 & 14.44
			& 19.00 & 32.43 & 11.33
			& 17.78 & 30.74 & 12.62 \\
			
			\textbf{FFN-only}
			& 17.57 & 30.97 & 11.66
			& 19.83 & 34.33 & 15.12
			& 19.47 & 33.39 & 11.86
			& 18.08 & 31.39 & 13.31 \\
			
			\midrule
			
			\textbf{w/o relation} 
			& 16.42 & 28.13 & 10.74
			& 19.43 & 33.04 & 14.41
			& 19.33 & 35.85 & 11.71
			& 17.59 & 30.82 & 12.74 \\
			
			\textbf{s-d only}
			& 16.44 & 28.24 & 10.71
			& 19.59 & 33.56 & 14.68
			& 19.02 & 34.17 & 11.44
			& 17.53 & 30.59 & 12.80 \\
			
			\midrule
			
			\textbf{w/o mean pooling}
			& 16.89 & 29.06 & 10.89
			& 19.89 & 33.95 & 14.88
			& 18.71 & 31.71 & 11.13
			& 17.47 & 29.98 & 12.63 \\
			
			\textbf{w/ dual attention}
			& 17.38 & 30.26 & 11.31
			& 20.37 & 34.56 & 15.02
			& 19.16 & 33.12 & 11.83
			& 17.90 & 30.89 & 12.93 \\
			
			\midrule
			
			\textbf{\texttt{LoReST}}
			& \textbf{16.18} & \textbf{27.85} & \textbf{10.70}
			& \textbf{19.41} & \textbf{32.98} & \textbf{14.23}
			& \textbf{18.32} & \textbf{31.32} & \textbf{10.90}
			& \textbf{17.12} & \textbf{29.47} & \textbf{12.19} \\
			
			\bottomrule
		\end{tabular}%
	}
\end{table*}

\subsection{Hyperparameter Study(\textbf{RQ3})}

To analyze the sensitivity of \texttt{LoReST} to key hyperparameters, we examine the effects of the number of neighbors \(k\), the number of regions \(R\), and the number of spatial encoder layers \(M\).

\begin{figure}[!t]
	\centering
	\includegraphics[width=0.5\textwidth]{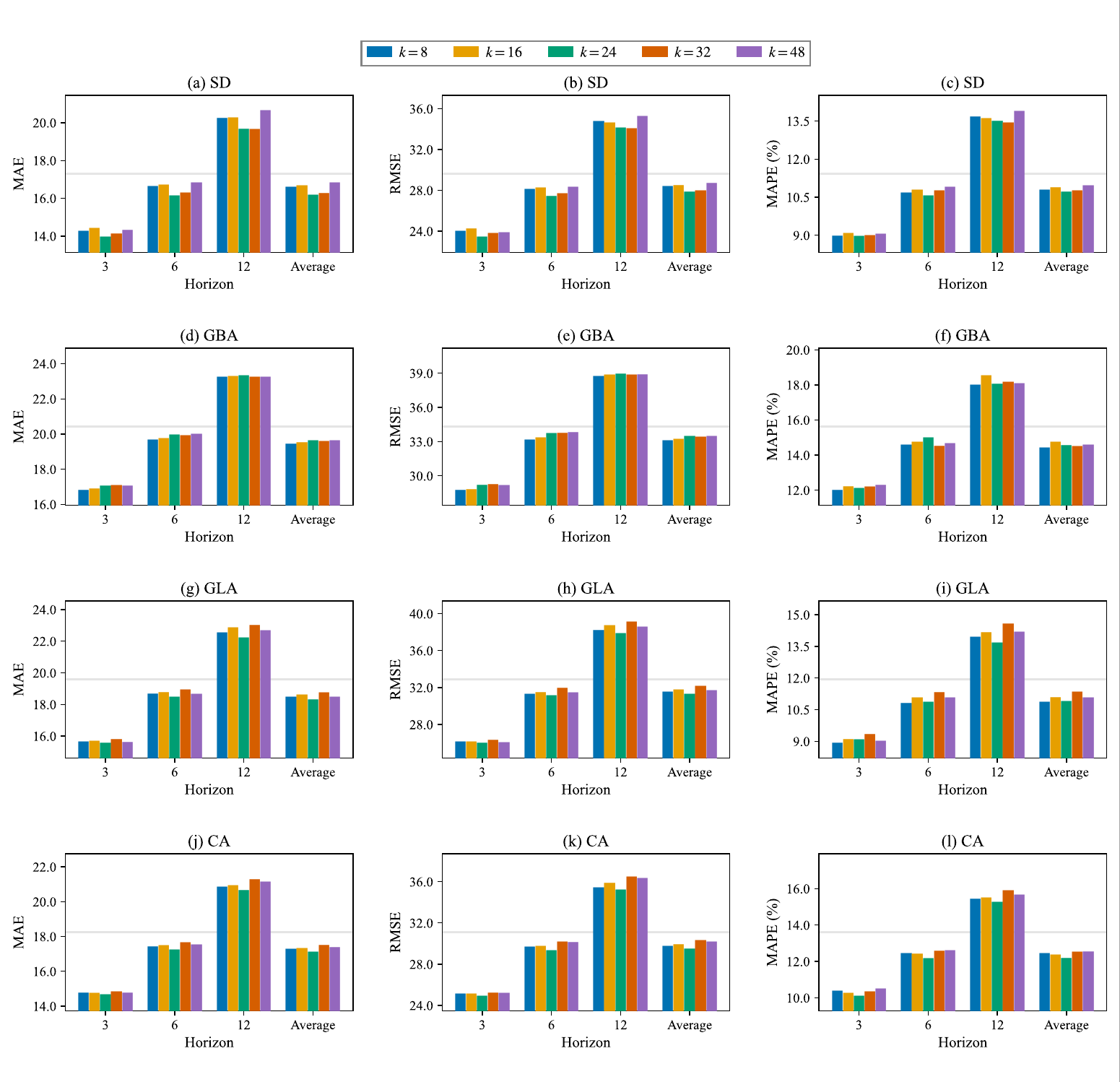}
	\caption{Hyperparameter study on number of neighbors $k$.}
	\label{fig: hyperk}
\end{figure}

To investigate the impact of \textbf{the number of neighbors} on model performance, we fix the number of regions at $r=16$ and the number of spatial encoder layers at $M=4$, and vary $k$ over $[8,16,24,32,48]$. The results are shown in the Fig. \ref{fig: hyperk}. Smaller neighborhoods may limit the coverage of local information, whereas further increasing the number of neighbors yields no consistent improvements. This suggests that local modeling requires a balance between information coverage and effective aggregation.

\begin{figure}[!t]
	\centering
	\includegraphics[width=0.5\textwidth]{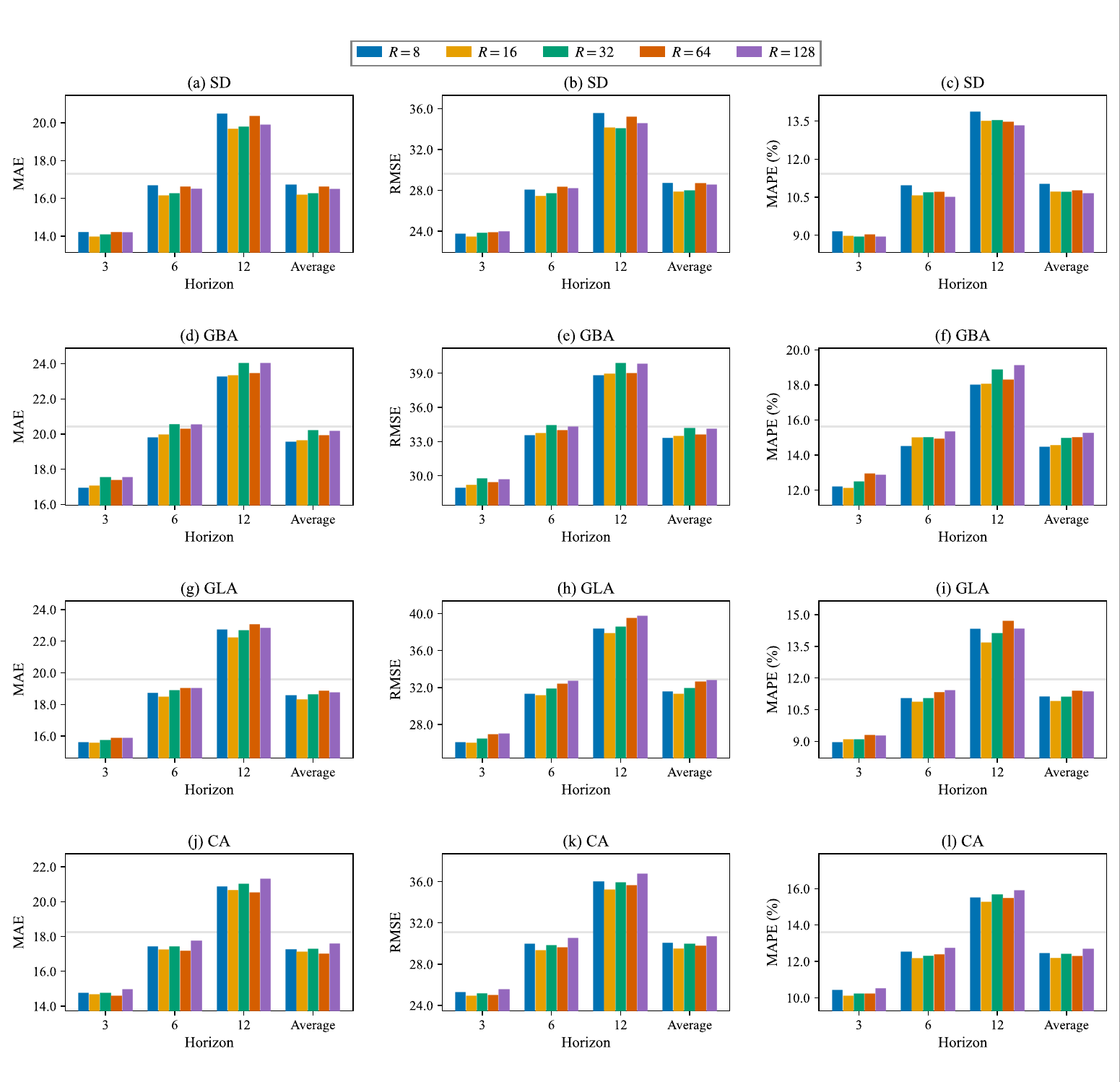}
	\caption{Hyperparameter studyon number of regions $r$.}
	\label{fig: hyperr}
\end{figure}

To investigate the impact of \textbf{the number of regions}, we fix the number of neighbors at $k=24$ and the number of spatial encoder layers at $M=4$, and vary $r$ over $[8,16,32,64,128]$. The results are shown in the Fig. \ref{fig: hyperr}. These results indicate that finer spatial partitioning does not necessarily improve overall forecasting performance, and that an appropriate region granularity facilitates the construction of effective cross-region context.

\begin{figure}[!t]
	\centering
	\includegraphics[width=0.5\textwidth]{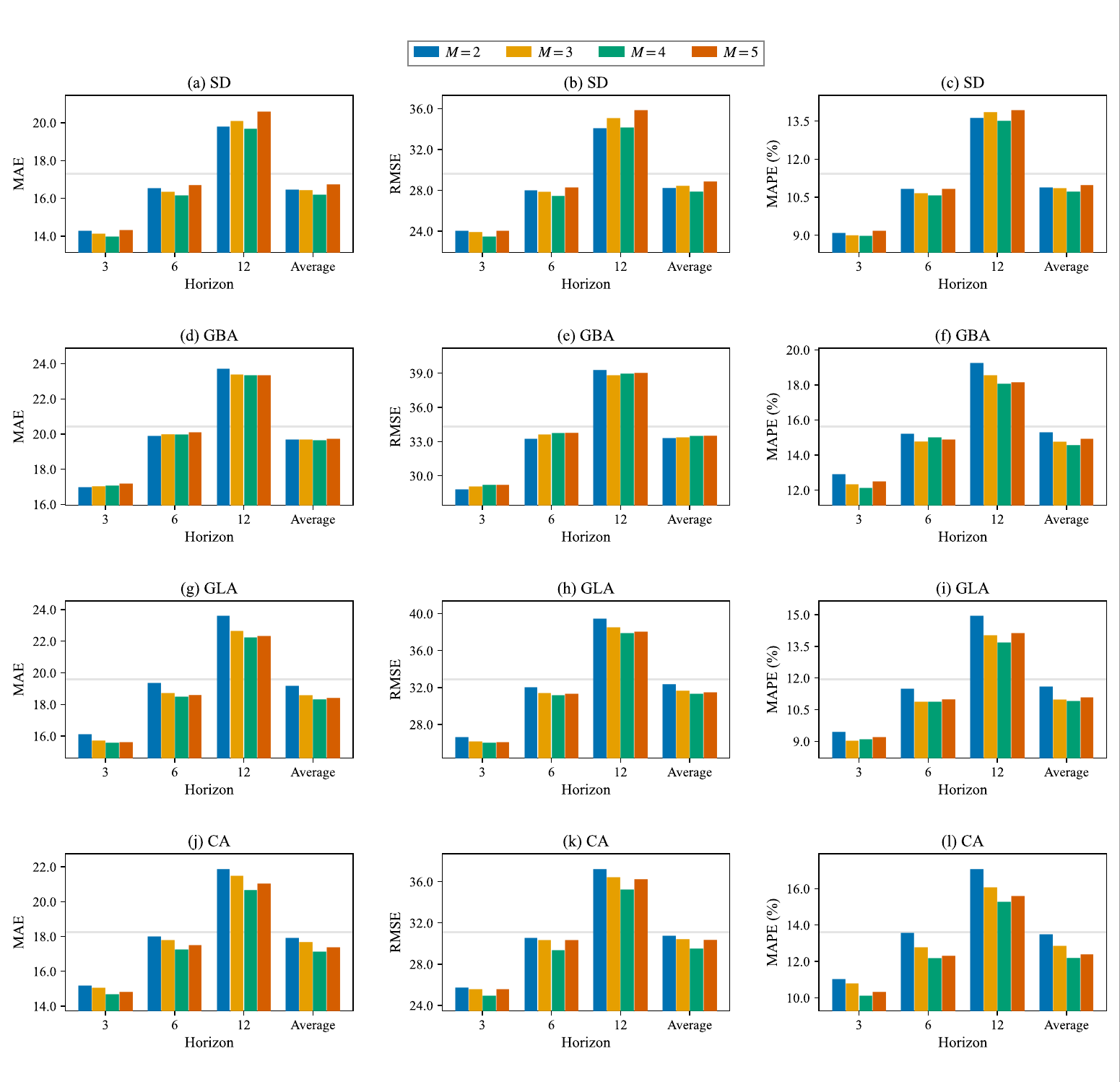}
	\caption{Hyperparameter study of Spatial Encoder Layers $M$.}
	\label{fig: hyperm}
\end{figure}

To investigate the impact of \textbf{model capacity}, we fix the number of neighbors at $k=24$ and the number of regions at $r=16$, and vary the number of layers $M$ over $[2,3,4,5]$. The results are shown in the Fig. \ref{fig: hyperm}. As can be observed, the model performance begins to degrade when the number of layers reaches 5. This suggests that, with spatial knowledge fully exploited, effective forecasting can be achieved without stacking excessive model capacity.

\subsection{Case Study}

\subsubsection{Analysis of Local Heterogeneity(\textbf{RQ4})}

\begin{figure}[!t]
	\centering
	\includegraphics[width=0.5\textwidth]{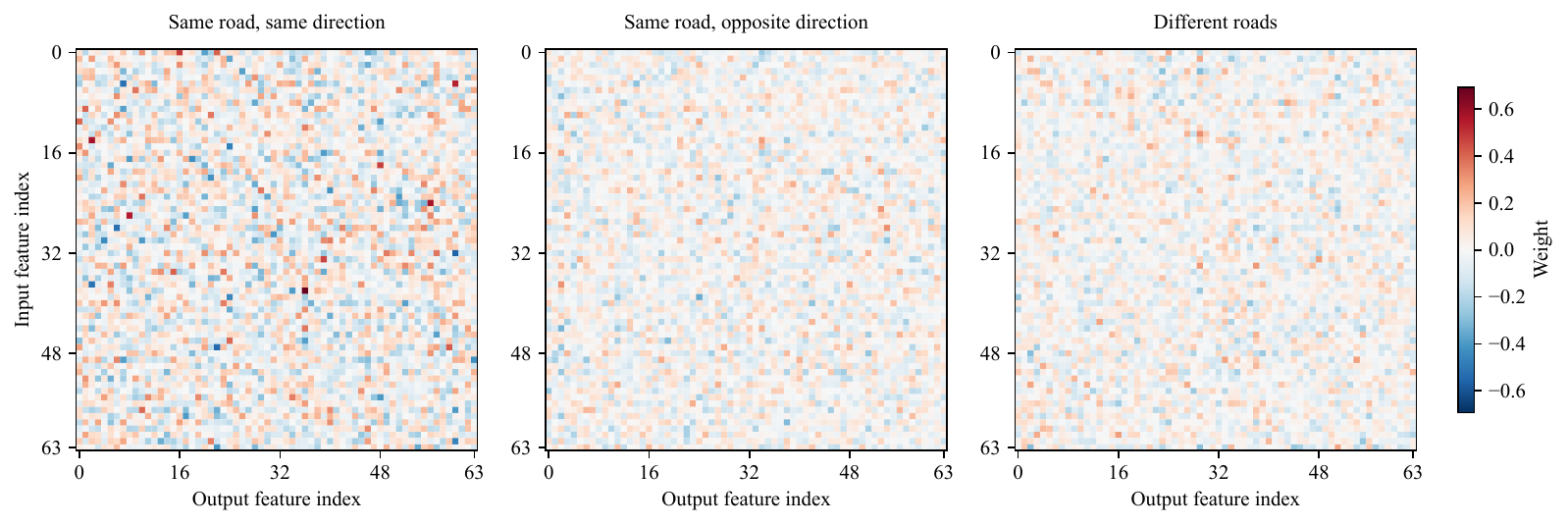}
	\caption{Learned relation-specific transformation matrices in the last spatial encoder layer on GLA. From left to right: same-road, same-direction; same-road, opposite-direction; and different-road relations. All heatmaps share the same color scale, with red and blue indicating positive and negative weights, respectively.}
	\label{fig: relaweights}
\end{figure}

To investigate whether \texttt{LoReST} learns local heterogeneity, we visualize the three relation-specific transformation matrices. As shown in the Fig. \ref{fig: relaweights}, the matrices exhibit distinct weight distributions. In particular, the matrix associated with same-road, same-direction neighbors displays more pronounced positive and negative weights, whereas the other two matrices exhibit smaller variations in weight magnitude. This indicates that the model learns differentiated feature transformations for different road relations rather than applying a uniform message mapping, providing a parameter-level illustration of how relation-aware local aggregation captures local heterogeneity.

\begin{figure}[!t]
	\centering
	\includegraphics[width=0.5\textwidth]{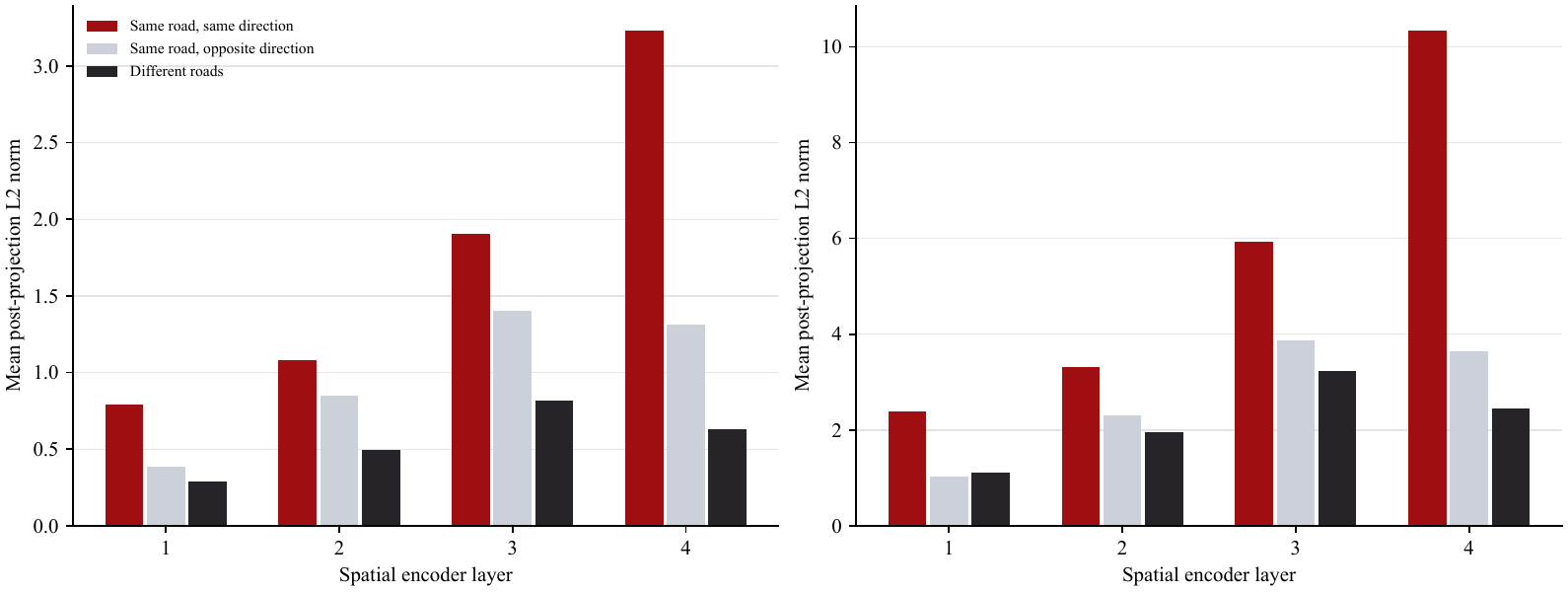}
	\caption{Message magnitudes for three road-relation types across spatial encoder layers. Left: mean L2 norms of relation-specific messages over all nodes. Right: mean L2 norms after relation-group-size normalization, evaluated only on nodes with all three neighbor types.}
	\label{fig: weightmessage}
\end{figure}

To further examine how the model utilizes information from different road relations, we compute the mean \(L_2\) norm of each relation-specific message after output projection at each spatial encoder layer. As shown in the Fig. \ref{fig: weightmessage}, messages from same-road, same-direction neighbors consistently exhibit greater magnitudes than those from the other two categories, with more pronounced differences in deeper layers. To reduce the influence of unequal neighbor counts, we further conduct a group-size-normalized comparison on nodes containing all three relation types. The message strengths of the three transformation matrices follow the same ordering as the correlations shown in Fig. \ref{fig: relations}, indicating that the relation-aware local aggregation module has successfully learned local heterogeneity.

\subsubsection{Prediction Visualization(\textbf{RQ5})}

\begin{figure}[!t]
	\centering
	
	\subfloat[3-step prediction (45 min).]{%
		\includegraphics[width=\linewidth]{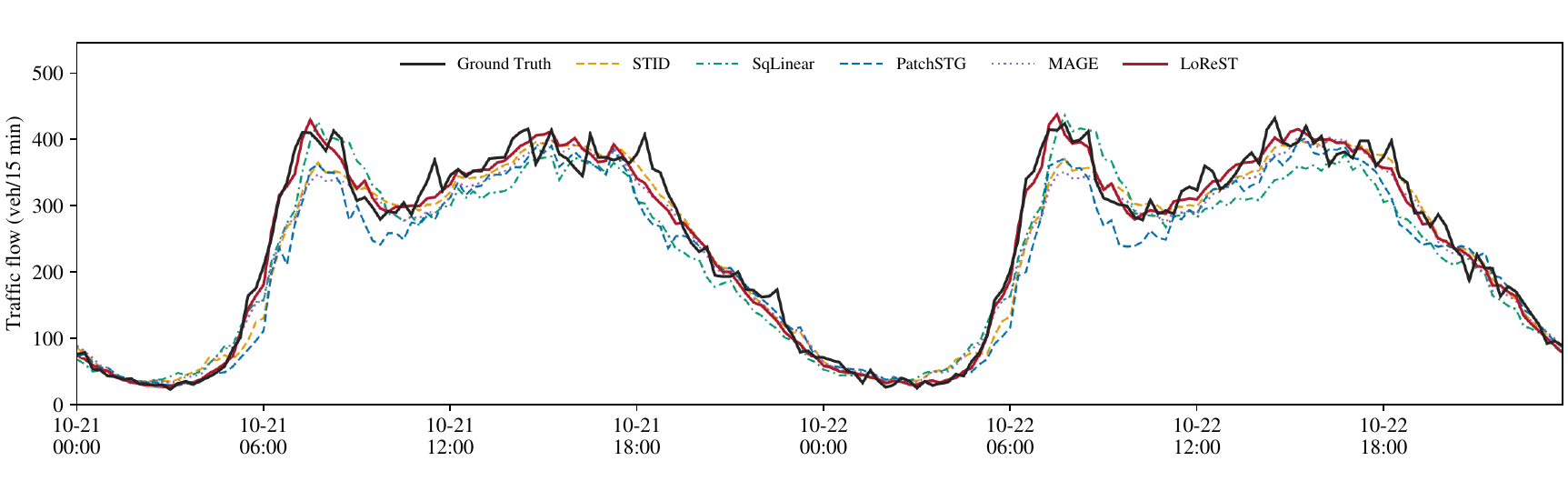}%
		\label{fig:prediction_h3}
	}
	
	\par\vspace{2pt}
	
	\subfloat[6-step prediction (90 min).]{%
		\includegraphics[width=\linewidth]{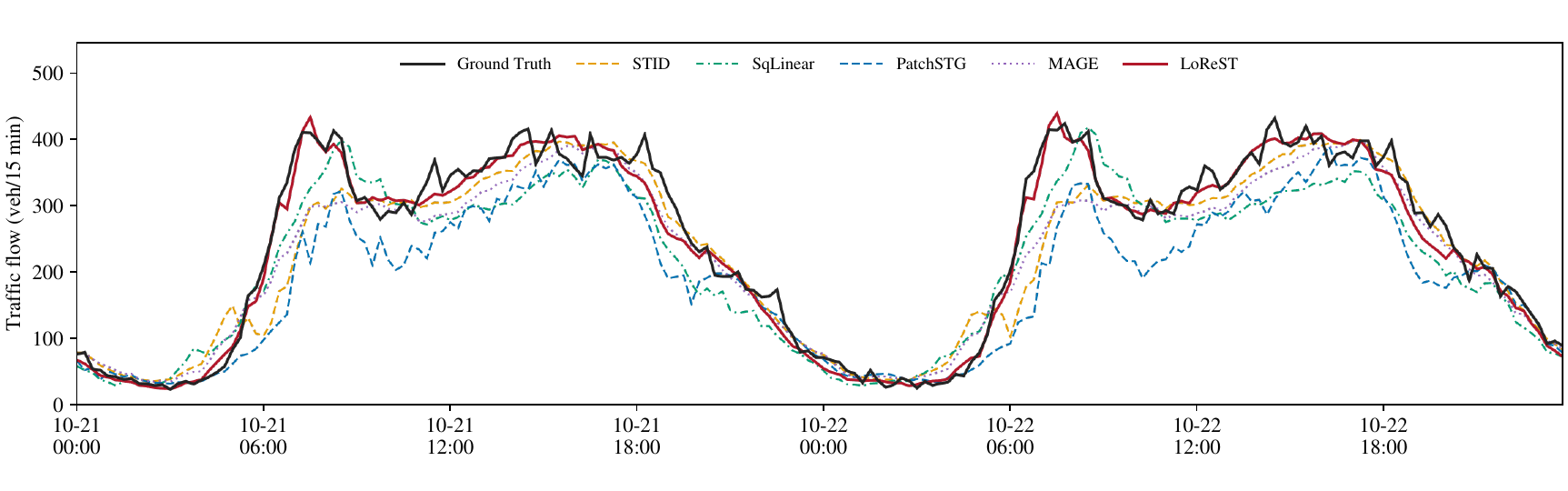}%
		\label{fig:prediction_h6}
	}
	
	\par\vspace{2pt}
	
	\subfloat[12-step prediction (180 min).]{%
		\includegraphics[width=\linewidth]{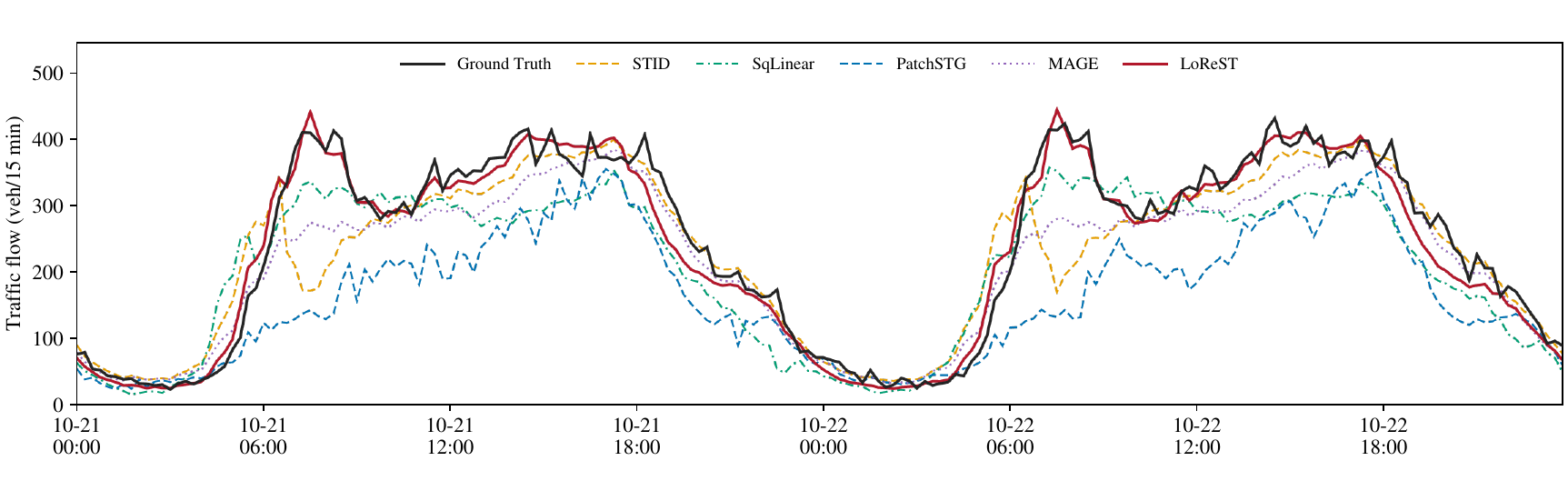}%
		\label{fig:prediction_h12}
	}
	
	\caption{An illustrative forecasting case on CA at three prediction horizons.}
	\label{fig:prediction_comparison}
\end{figure}

To visually compare the forecasting performance of different models, we present their predictions over the same time interval from the CA at forecasting horizons of 3, 6, and 12 steps. As shown in Fig. \ref{fig:prediction_comparison}, all models generally capture the periodic patterns of traffic flow, but differ in their predictions during peak periods, rapid transitions, and local fluctuations. In comparison, \texttt{LoReST}’s predictions align more closely with the ground truth, better capturing peak magnitudes and changes during rising and falling phases. As the forecasting horizon increases, capturing local variations becomes more challenging for all models, while \texttt{LoReST} maintains good trend-tracking performance.

\section{Conclusion and Future Work}\label{sec: conclusion}

To address complex spatial dependencies in large-scale road networks, this paper proposes \texttt{LoReST}, a local–region joint learning framework for traffic flow forecasting. By jointly modeling node-level local dependencies and region-level long-range context, \texttt{LoReST} learns effective spatial representations for node-level forecasting. It incorporates spatial prior knowledge, including geographic locations, road identities, and travel directions, to construct candidate neighborhoods and partition the road network into regions. At the local level, the relation-aware local aggregation module applies distinct feature transformations to neighbor information under different road relationships, capturing heterogeneous local spatial dependencies. At the region level, the cross-region spatial interaction module constructs regional representations through mean pooling and exchanges long-range context through inter-region attention, enriching node representations with information beyond their local neighborhoods. Together, these mechanisms preserve fine-grained local information while avoiding the high computational cost of all-node interactions. Experiments on the four datasets of the LargeST benchmark demonstrate that \texttt{LoReST} achieves state-of-the-art forecasting performance. Ablation studies further validate the effectiveness of local relation modeling and cross-region context interaction.

In future work, we will explore how to incorporate spatial prior knowledge into pretrained time-series foundation models, combining transferable local relation modeling with cross-region context interaction to enhance zero-shot forecasting on unseen road networks.


\bibliographystyle{IEEEtran}
\bibliography{IEEEabrv, ref}

\vfill
\end{document}